\documentclass[journal]{IEEEtran}
\usepackage{amsmath,amsfonts}
\usepackage{array}
\usepackage[caption=false,font=footnotesize,labelfont=rm,textfont=rm]{subfig}
\usepackage{textcomp}
\usepackage{stfloats}
\usepackage{url}
\usepackage{verbatim}
\usepackage{graphicx}
\usepackage{cite}
\usepackage{color}
\usepackage{amssymb}
\usepackage{booktabs}
\usepackage{multirow} 
\usepackage{makecell}
\usepackage{threeparttable}

\usepackage[ruled,norelsize,linesnumbered]{algorithm2e} 
\usepackage[colorlinks,linkcolor=blue,anchorcolor=blue,citecolor=blue]{hyperref}

\usepackage{pifont} 

\begin{document}

\title{A Human-Oriented Cooperative Driving Approach: Integrating Driving Intention, State, and Conflict}

\author{Qin Wang,~Shanmin Pang,~\IEEEmembership{Member, IEEE}, ~Jianwu Fang,~\IEEEmembership{Member, IEEE},~Shengye Dong,~Fuhao Liu,\\Jianru Xue,~\IEEEmembership{Member, IEEE},  Chen Lv,~\IEEEmembership{Senior Member, IEEE}
\thanks{This manuscript was supported  by the National Key R\&D Program of China (Grant No. 2022ZD0117903) and the Outstanding Youth Foundation of Shaanxi Province (Grant No. 2025JC-JCQN-092).}
\thanks{Qin Wang, Shanmin Pang, Shengye Dong and Fuhao Liu are with the School of Software Engineering, Xi'an Jiaotong University, Xi'an, China (wqin98@163.com, pangsm@xjtu.edu.cn, dongshengye6@163.com, liufuhao@stu.xjtu.edu.cn).}
\thanks{Jianwu Fang and Jianru Xue are with the School of Artificial Intelligence, Xi'an Jiaotong University, Xi'an, China (fangjianwu, jrxue@xjtu.edu.cn).}
\thanks{Chen Lv is with the School of Mechanical and Aerospace Engineering, Nanyang Technological University, Singapore (lyuchen@ntu.edu.sg).}}



\maketitle

\begin{abstract}
Human-vehicle cooperative driving serves as a vital bridge to fully autonomous driving by improving driving flexibility and gradually building driver trust and acceptance of autonomous technology. To establish more natural and effective human-vehicle interaction, we propose a Human-Oriented Cooperative Driving (HOCD) approach that primarily minimizes human-machine conflict by prioritizing driver intention and state. In implementation, we take both tactical and operational levels into account to ensure seamless human-vehicle cooperation. At the tactical level, we design an intention-aware trajectory planning method, using intention consistency cost as the core metric to evaluate the trajectory and align it with driver intention. At the operational level, we develop a control authority allocation strategy based on reinforcement learning, optimizing the policy through a designed reward function to achieve consistency between driver state and authority allocation. The results of simulation and human-in-the-loop experiments demonstrate that our proposed approach not only aligns with driver intention in trajectory planning but also ensures a reasonable authority allocation. Compared to other cooperative driving approaches, the proposed HOCD approach significantly enhances driving performance and mitigates human-machine conflict.
The code is available at https://github.com/i-Qin/HOCD.

\end{abstract}


\section{Introduction}
\IEEEPARstart{R}{ecently} remarkable progress in mobile communication, sensor technology, and artificial intelligence, collectively contribute to the significant development of autonomous driving. Nevertheless, there are still various issues to be solved, such as complex regulatory landscape, public safety concerns, insurance issues, and potential legal liabilities, etc., which lead to a yawning gulf between reality and fully autonomous driving \cite{sever2024automated}. Therefore, human-vehicle cooperation driving still plays a pivotal role in the transition from manual to fully autonomous driving \cite{zhang2024human}. By retaining the human driver actively engaged in the system loop, the human-vehicle cooperative driving can take full advantage of the human experience in situational awareness and ensure safety and adaptability \cite{tan2021human, 3muzahid2024survey}. However, how to achieve efficient and flexible cooperation with automation has attracted significant attention \cite{peintner2024design}.

An effective approach to achieving human-vehicle cooperative driving is shared control, a concept derived from human-machine cooperation, where the human and automation work together simultaneously on the same task \cite{5marcano2020review},\cite{6sheridan1978human}. Shared control contains two types: direct shared control and indirect shared control \cite{8wang2020decision}. Direct shared control typically involves the haptic guidance system, which provides force feedback and interacts with the human driver via haptic signals \cite{14abbink2010neuromuscular, 15mars2014analysis, 16abbink2012haptic}. With the development of the steer-by-wire technique, more flexible indirect shared control methods have been proposed, where interaction occurs virtually through the combination of the driver command input and the controller desired input. 
In essence, the key to shared control lies in designing a reasonable authority allocation strategy between the human driver and automation to adapt to ever-changing driving circumstances \cite{7yang2021review}. This can be achieved through approaches such as function-based \cite{18jiang2017shared,12fang2024human,19li2021shared, 9fang2023human, 20nguyen2017sensor}, 
rule-based \cite{10lu2022shared,17li2018shared}, control theory-based\cite{23_2022adaptive,24liu2022human}, and game-based \cite{25dai2023bargaining,26guo2023game}. While these methods demonstrate strong performance, they often lack flexibility in integrating multi-objective strategies.
In contrast, Reinforcement Learning (RL) provides a promising alternative by facilitating continuous interaction with the environment and effectively balancing multiple objectives without relying on precise models, making it well-suited for dynamic and complex scenarios\cite{rl-wang2024human, rl-xie2022coordination, 48yan2024human}.

Furthermore, since humans play a pivotal role in human-machine cooperation, driver-related factors are increasingly recognized as key components in the design of authority allocation strategy. Consequently, recent studies have incorporated various driver characteristics, such as driving style, intention, state, and ability, to improve the effectiveness of these systems. 
For example, Fang \textit{ et al.}~\cite{9fang2023human} used driving intent and ability to represent driver characteristics. 
Priority was given to enhancing cooperation between automation and different types of drivers, such as aggressive, typical, or conservative drivers \cite{10lu2022shared}. 
Liu \textit{et al.}~\cite{11fang2023servo} proposed an authority allocation method by assessing driver state and long-term ability. A time-varying driver steering model considering fatigue driving is established to address the model mismatch \cite{12fang2024human}. 

\begin{figure}[!t]
\centering
\includegraphics{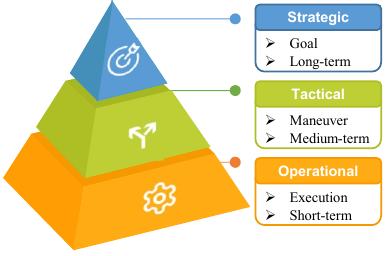}  
\caption{Three levels of human-vehicle cooperative driving.} 
\label{fig1}
\end{figure}

Although these approaches incorporate driver characteristics and mitigate conflicts between the driver and automation to some extent, their effectiveness remains mainly limited to the control level. As illustrated in Fig.~\ref{fig1}, human-vehicle cooperative driving operates on three interdependent levels \cite{3muzahid2024survey},\cite{8wang2020decision}: strategic, tactical, and operational. The strategic level focuses on long-term objectives, the tactical level focuses on real-time driver maneuvers, such as obstacle avoidance and lane changes, and the operational level refers to immediate execution relating to vehicle control. Only through seamless cooperation across the three levels can human factors and system capabilities be effectively addressed. Therefore, it is essential to consider the tactical level alongside the operational level. If trajectory planning at the tactical level is not addressed properly, conflicts may arise even when the driver behaves correctly \cite{5marcano2020review}. Currently, most trajectory planning approaches focus on generating collision-free trajectories based on the surrounding environment and vehicle dynamics. However, human-vehicle cooperation has yet to be considered in trajectory planning.

Building on the above analysis, we propose a human-oriented human-vehicle cooperative driving approach utilizing reinforcement learning. Our approach integrates human cognitive flexibility with the precision and reliability of automation for overall system optimization. In particular, we apply the principle of human-oriented automation to minimize human-machine conflicts from two aspects. Firstly, recognizing that conflicts can stem from the mismatch between human expectations and automation in the higher-level task of trajectory planning, we plan trajectories guided by human driver intention at the tactical level. This method enhances driver flexibility and helps mitigate potential conflicts. Secondly, conflicts can also arise from an inappropriate strategy for control authority allocation between the human and automation. The human-machine conflict increases when the automation fails to adjust according to the driver's current state. To address this, we fully leverage the advantages of reinforcement learning in solving complex decision-making problems and implement it at the operational level to dynamically allocate authority, taking driver state and potential conflicts into account.

The contributions of this manuscript are as follows:
\begin{itemize}
    \item We propose a human-oriented human-vehicle cooperative driving approach that integrates driver intention and state, aiming to reduce potential conflicts at both the tactical and operational levels.
    
    \item We design an intention-aware trajectory planning method that constructs lateral and longitudinal trajectory models in the Frenet frame, and generates and checks a set of trajectories, with intention consistency serving as the key cost function to select the optimal trajectory.
    
    \item We develop a novel RL-based authority allocation strategy that incorporates driver state as input and utilizes a multi-objective reward function to minimize conflicts between driver state and authority allocation.
\end{itemize}

 
 \section{Related Work}

 \subsection{Trajectory Planning}
Methods for trajectory planning are typically classified into four categories: sampling-based, geometry-based, optimization-based, and graph-based approaches.

Sampling-based planning involves sampling a set of discrete points in the state space and constructing connections among the points to generate candidate paths \cite{27elbanhawi2014sampling}. The most common sampling-based algorithms are Probabilistic Roadmap (PRM) \cite{28kavraki1996probabilistic} and Rapidly Exploring Random Tree (RRT) \cite{29lavalle2001randomized}. PRM constructs a graph of potential paths within a given map by distinguishing between free and occupied spaces, followed by applying a graph search algorithm to identify a viable path. Although effective, PRM does not always guarantee an optimal solution. In contrast, RRT is designed to efficiently explore nonconvex, high-dimensional spaces by incrementally building a space-filling tree through random sampling.

Geometry-based planning generates paths with high continuity and smoothness by fitting curves through key nodes \cite{33song2023review}. Commonly used interpolating curves include polynomial curves \cite{34lee2012unified}, Dubins curves \cite{35yang20132d}, Bezier curves \cite{36qian2016motion}, and spline curves \cite{37maekawa2010curvature}. Due to low computational cost and high real-time performance, geometry-based methods are often combined with other planning techniques for enhanced efficiency.

Optimization-based planning converts trajectory planning to a multi-objective optimization problem subject to a set of constraints and cost functions \cite{30lim2018hierarchical}. Li et al. \cite{31li2022autonomous} applied this method in the Cartesian frame within the continuous solution space to find optimal trajectories, proposing an iterative computation framework to address nonconvex kinematic constraints. Guo et al. \cite{32guo2022down} simplified the optimization problem by optimizing critical variables, thereby enhancing the efficiency of the optimization-based trajectory planner.

Graph-based planning discretizes the motion state space, constructs a graph structure, and then finds the optimal path using the cost function. Dijkstra's algorithm \cite{38hwang2003fast}, A* \cite{39hart1968formal}, Hybrid A* \cite{40kurzer2016path}, and D* \cite{41stentz1994optimal} are widely used graph search algorithms.

\begin{figure*}[!t]
\centering
\includegraphics[width=0.97\textwidth]{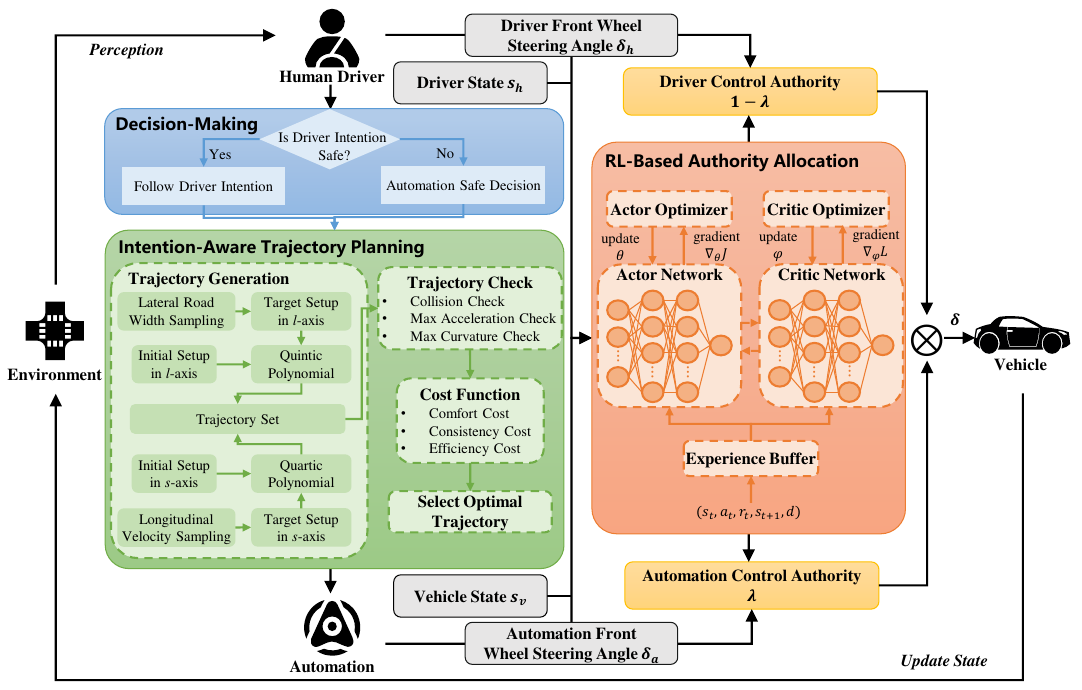}
\caption{Framework of the proposed human-oriented cooperative driving approach.}
\label{fig2}
\vskip -0.1in
\end{figure*}

 \subsection{Authority Allocation Strategies}
In the literature, studies on authority allocation can generally be classified into five categories: function-based, rule-based, control theory-based, game-based, and RL-based approaches.

Function-based allocation includes three types of function: piecewise function, exponential function, and U-shape function. Jiang \textit{et al.} \cite{18jiang2017shared} designed a three-level piecewise function based on safe, hysteresis, and dangerous subsets, which are segmented according to human inputs and constraints. Fang \textit{et al.}\cite{12fang2024human} defined a piecewise function based on varying fatigue level ranges. The exponential function was employed to integrate driver behavioral parameters and states, providing a weighted approach to shared control \cite{19li2021shared}. Additionally, driving ability and involvement were incorporated into the exponential function to design an authority allocation strategy \cite{9fang2023human}. Nguyen \textit{et al.} \cite{20nguyen2017sensor} used a U-shape function to calculate the driver requirement for assistance based on driving workload.

Rule-based allocation utilizes fuzzy logic to model the relationship between various influencing factors and shared control authority. Lu \textit{et al.} \cite{10lu2022shared} formulated fuzzy rules based on the time-to-collision (TTC) and driver steering characteristics, such as the normalized steering angle input. A fuzzy-rule-based method was also implemented to calculate the cooperative coefficient, taking into account driver intention and situational assessment \cite{17li2018shared}.

Control theory-based allocation dynamically determines authority levels using Model Predictive Control (MPC). In \cite{23_2022adaptive}, the reference authority was calculated by the reference model and then was provided to the MPC to determine real-time authority level according to the vehicle state and the driver characteristic coefficient. Liu \textit{et al.} \cite{24liu2022human} developed an online authority optimization strategy using constrained MPC, accounting for the driver driving skill and intention. 

Game-based allocation models the decision-making of the human driver and automation by game theory to calculate the authority. Given strict game theory assumptions, demands for the driver's desired trajectory, and high requirements for controllers, Dai \textit{et al.} \cite{25dai2023bargaining} proposed an allocation strategy based on the bargaining game for steer-by-wire vehicles. This approach decouples the human-machine structure and increases the flexibility of game modeling. To address the computational limitations associated with strong numerical approximations, the game equilibrium strategy was derived using a piecewise affine method, and then the Nash equilibrium solution was obtained through a convex iteration method \cite{26guo2023game}.

RL-based allocation learns the optimal policy by receiving feedback during interactions with the environment. Compared to the aforementioned methods, reinforcement learning demonstrates better adaptability, as it can dynamically adjust according to changes and feedback from the environment, enabling it to make decisions in uncertain and complex environments.
Xie \textit{et al.} \cite{rl-xie2022coordination} pioneered the application of reinforcement learning in human-vehicle cooperative steering to achieve optimal path-following. Wang \textit{et al.} \cite{rl-wang2024human} improved the driver preview model to output the driver steering angle, and incorporate it as one of the input states for reinforcement learning. Similarly, Yan \textit{et al.} \cite{48yan2024human} utilized reinforcement learning to address reference-free shared driving problems. However, these approaches overlook driver-related factors in control authority allocation during reinforcement learning training, potentially leading to human-machine conflicts.

\section{Methodology}

The main framework of the Human-Oriented Cooperative Driving (HOCD) approach is illustrated in Fig.~\ref{fig2}. The approach consists of intention-aware trajectory planning and an RL-based authority allocation strategy. Initially, the human driver may express a lane change intention at any time while perceiving the traffic environment. Upon making a decision, such as activating the turn signal, the automation evaluates the safety of executing the driver's intention. If the intention is deemed safe, the automation plans a trajectory consistent with the driver's intention; otherwise, it plans a safe and optimal trajectory. 

Subsequently, the RL-based authority allocation strategy receives inputs including the driver control command, driver state, automation control command, and vehicle state. The reinforcement learning agent then optimizes the control authority dynamically during each control cycle. The final steering command is calculated by integrating the control inputs from both the driver and the automation using a linear weighted sum method, which is defined as follows:\begin{equation}
\label{linear weighed sum}
\delta = \lambda \delta_{a}+(1-\lambda)\delta_{h},
\end{equation}
where $\delta$ is the final front wheel steering angle,  $\delta_a$ is the automation front wheel steering angle, $\delta_h$ is the driver front wheel steering angle, and $\lambda \in[0,1]$ is the automation control authority. The authority $\lambda$ is dynamically adjusted in real-time to ensure seamless cooperation between the driver and the automation. As a special case, when $\lambda = 0$ the human driver has complete control over the vehicle without the assistance of automation. Conversely, when $\lambda = 1$, the automation takes full control, fully managing the vehicle operation. 


\subsection{Intention-Aware Trajectory Planning}
\subsubsection{Frenet Frame}

\begin{figure}[!t]
\centering
\includegraphics[width=0.85\columnwidth]{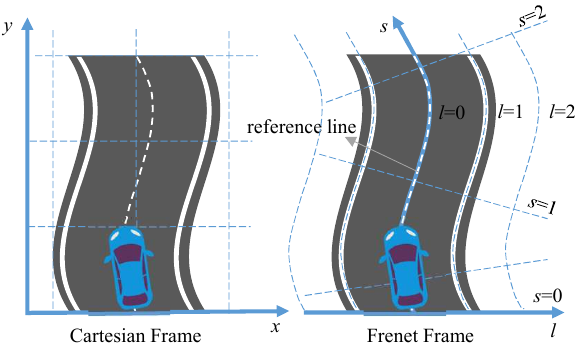}
\caption{Difference between the Frenet frame and the Cartesian frame.}
\label{fig3}
\vskip -0.1in
\end{figure}

Due to the curvature of most roads, it is challenging to calculate information such as driving distance, and lateral offset from lane centerline in a Cartesian frame. Developing a uniform and concise representation of road geometry is crucial for improving the efficiency of trajectory planning. To this end, Werling \textit{et al.} \cite{55werling2010optimal} presented the Frenet frame, which simplifies the generation of optimal trajectories for street scenarios. As shown in Fig.~\ref{fig3}, the lane centerline serves as the reference line in the Frenet frame, and the vehicle  motion is decoupled into longitudinal motion along the centerline and lateral motion perpendicular to it. 
This decoupling, corresponding to longitudinal velocity  planning and lateral trajectory planning, significantly reduces computational complexity compared to direct trajectory planning in the Cartesian frame. As a result, the Frenet frame has been widely adopted in trajectory planning.
Formally, the vector $[s, \dot s, \ddot s, l, \dot l, \ddot l, l^{'}, l^{''} ]$ represents the vehicle state at any time $t$, where $s$ denotes the longitudinal displacement along the tangent direction of the reference line, and $l$ denotes the lateral displacement along the normal direction of the reference line. The variables $\dot s$ and $\ddot s$ represent the longitudinal velocity and the longitudinal acceleration, respectively, while $\dot l$ and $\ddot l$ denote the lateral velocity and lateral acceleration. Additionally, $l^{'}$ and $l^{''}$ correspond to the first and second derivatives of $l$ with respect to $s$, respectively.

\subsubsection{Lateral and Longitudinal Motion Trajectory Planning}
Lateral motion trajectory planning refers to planning in the direction perpendicular to the vehicle motion, which involves tasks such as obstacle avoidance and lane changes, shaping the overall trajectory. The lateral trajectory $l(t)$ is modeled using a quintic polynomial: 
\begin{equation}
\label{eq2}
l(t)=c_{l0}+c_{l1}t+c_{l2}t^2+c_{l3}t^3+c_{l4}t^4+c_{l5}t^5.
\end{equation}

The lateral trajectory $l(t)$, along with its first and second derivative with respect to $t$, can be represented in matrix form as follows:
\begin{equation}
     \label{eq5}
     \begin{aligned}
    \begin{bmatrix}l(t)\\ \dot l(t)\\ \ddot l(t)\end{bmatrix}
    &=\begin{bmatrix}1&t&t^2\\0&1&2t\\0&0&2\end{bmatrix}\begin{bmatrix}c_{l0}\\c_{l1}\\c_{l2}\end{bmatrix}+\begin{bmatrix}t^3&t^4&t^5\\3t^2&4t^3&5t^4\\6t&12t^2&20t^3\end{bmatrix}\begin{bmatrix}c_{l3}\\c_{l4}\\c_{l5}\end{bmatrix}\\
    &=M_{d1}(t)\cdot\begin{bmatrix}c_{l0}\\c_{l1}\\c_{l2}\end{bmatrix}+M_{d2}(t)\cdot\begin{bmatrix}c_{l3}\\c_{l4}\\c_{l5}\end{bmatrix}.
    \end{aligned}
\end{equation}


The initial state at time $t_0$ is defined as $l_0=[l(t_0), \dot l(t_0), \ddot l(t_0)]$, while the target state at time $t_i$ is $l_i=[l(t_i), \dot l(t_i), \ddot l(t_i)]$. To simplify calculations, the initial time is set to $t_0=0$. As such, the coefficients $c_{l0}$, $c_{l1}$, and $c_{l2}$ can be directly computed as: $c_{l0}=l(0)$, $c_{l1}=\dot l(0)$,  $c_{l2}=\ddot l(0)/2$.

Let the target time be $t_i=\tau~(\tau>0)$. The polynomial coefficients $c_{l3}$, $c_{l4}$, and $c_{l5}$ can be calculated as:
\begin{equation}
    \begin{bmatrix}c_{l3}\\c_{l4}\\c_{l5}\end{bmatrix}=M_{d2}{}^{-1}(\tau)\cdot\begin{pmatrix}\begin{bmatrix}l(\tau)\\ \dot l(\tau)\\ \ddot l(\tau)\end{bmatrix}-M_{d1}(\tau)\cdot\begin{bmatrix}l(0)\\ \dot l(0)\\\frac{\ddot{l}(0)}{2}\end{bmatrix}\end{pmatrix}.
\end{equation}


Longitudinal motion trajectory planning focuses on velocity planning along a predetermined path, which determines the vehicle motion throughout the entire trajectory. In this study, the vehicle longitudinal velocity is assumed to remain constant, allowing the target position to be disregarded. Consequently, the longitudinal trajectory $s(t)$ can be modeled using a quartic polynomial:
\begin{equation}
\label{eq7}
s(t)=c_{s0}+c_{s1}t+c_{s2}t^2+c_{s3}t^3+c_{s4}t^4.
\end{equation}

The first and second derivatives of $s(t)$ with respect to $t$ can be represented in matrix form as follows: 
\begin{equation}
\label{eq10}
    \begin{aligned}
        \begin{bmatrix}\dot{s}(t)\\\ddot{s}(t)\end{bmatrix}&=\begin{bmatrix}1&2t\\0&2\end{bmatrix}\begin{bmatrix}c_{s1}\\c_{s2}\end{bmatrix}+\begin{bmatrix}3t^2&4t^3\\6t&12t^2\end{bmatrix}\begin{bmatrix}c_{s3}\\c_{s4}\end{bmatrix}\\
        &=M_{s1}(t)\cdot\begin{bmatrix}c_{s1}\\c_{s2}\end{bmatrix}+M_{s2}(t)\cdot\begin{bmatrix}c_{s3}\\c_{s4}\end{bmatrix}.
    \end{aligned}
\end{equation}


The initial state at time $t_0$ is defined as $s_0=[s(t_0), \dot s(t_0), \ddot s(t_0)]$, and the target state at time $t_i$ is $s_1=[\dot s(t_i), \ddot s(t_i)]$. When $t_0=0$, the coefficients $c_{s0}$, $c_{s1}$, and $c_{s2}$  can be directly calculated as: $c_{s0}=s(0)$, $c_{s1}=\dot s(0)$, and $c_{s2}=\ddot s(0)/2$.

Given that the target time is set to $t_i=\tau$, the polynomial coefficients $c_{s3}$ and $c_{s4}$ can be determined as follows:
\begin{equation}
    \begin{bmatrix}c_{s3}\\c_{s4}\end{bmatrix}=M_{s1}(\tau)^{-1}\cdot\left(\begin{bmatrix}\dot{s}(\tau)\\\ddot{s}(\tau)\end{bmatrix}-M_{s1}(\tau)\cdot\begin{bmatrix}\dot{s}(0)\\\frac{\ddot{s}(0)}{2}\end{bmatrix}\right).
\end{equation}

\subsubsection{Candidate Trajectory Check}
To improve the calculation efficiency, trajectories that fail to meet the specified constraints are filtered out through the candidate trajectory check in the Cartesian frame. The process includes checks for curvature, acceleration, and collisions.

Specifically, for the curvature check, the trajectory curvature must not exceed the maximum allowable value at any point along the trajectory: 
\begin{equation}
\label{eq12}
\kappa[i] \leqslant \kappa_{\max }, 0 \leqslant i \leqslant N,
\end{equation}
where $\kappa_{max}$ represents the maximum allowable curvature, $i$ denotes the $i$-th trajectory point, and $N$ represents the total number of trajectory points. 

Similarly, for the acceleration check, the trajectory acceleration must not exceed the maximum allowable acceleration at any point along the trajectory:
\begin{equation}
\label{eq13}
a[i] \leqslant a_{\max }, 0 \leqslant i \leqslant N,
\end{equation}
where $a_{max}$ represents the maximum allowable acceleration. 

For the collision check, the trajectory must not overlap with the obstacle collision area. This condition is mathematically expressed as:
\begin{equation}
\label{eq14}
(x_i-x_{obs} )^2+(y_i-y_{obs} )^2 < (r+r_{obs} )^2,0 \leqslant i \leqslant N,
\end{equation}
where $x_i$ and $y_i$ denote the horizontal and vertical coordinates of the $i$-th trajectory point in the Cartesian frame, $x_{obs}$ and $y_{obs}$ represent the horizontal and vertical coordinates of the obstacle in the Cartesian frame, and $r$ and $r_{obs}$ represent the radii of the circular bounding boxes of the ego vehicle and the obstacle, respectively.

\begin{algorithm}[t!]
\caption{Intention-Aware Trajectory Planning}\label{alg1}
\KwIn{human target lane $l_{target}$, road width $W$, predict time $T$, velocity $V$}
\KwOut{the optimal trajectory}
Initialize: state in $l$-axis $l_0=[l(t_0), \dot l(t_0), \ddot l(t_0)]$,  state in $s$-axis $s_0=[s(t_0), \dot s(t_0), \ddot s(t_0)]$\;
\For {$d_i \leftarrow W_{min}$ \KwTo $W_{max}$}{
    \For {$t_i \leftarrow T_{min}$ \KwTo $T_{max}$}{
        lateral trajectory $\leftarrow$ Quintic Polynomial $(l(t_0)$, $\dot l(t_0)$, $\ddot l(t_0)$, $d_i$, $\dot l(t_i)$, $\ddot l(t_i), t_i)$\;
        \For {$v_i \leftarrow V_{min}$ \KwTo $V_{max}$}{
            longitudinal trajectory $\leftarrow$ Quartic Polynomial $(s(t_0)$, $\dot s(t_0)$, $\ddot s(t_0)$, $v_i$, $\ddot s(t_i)$, $t_i)$\;
            Compute the cost function defined in Equation \eqref{eq15}\;
            Append trajectory and cost to the set\;
        }
    }
}
Check the trajectories according to Equations~\eqref{eq12},~\eqref{eq13} and ~\eqref{eq14}\;
Select the optimal trajectory with the minimum cost.
\end{algorithm}

\subsubsection{Cost Function}
To further determine the optimal trajectory, each filtered trajectory is evaluated using a specifically designed cost function. The trajectory with the lowest cost is then identified as the optimal choice. In this study, the cost function incorporates three key factors: comfort, consistency between human driver intention and automation, and efficiency.

The derivative of acceleration, referred to jerk, is a common cost for assessing trajectory comfort. Accordingly, the derivatives of lateral and longitudinal acceleration are incorporated into the cost function as follows:
\begin{equation}
    C_\mathrm{comfort}(t)=\int \dddot{l}^2(t)dt+\int \dddot{s}^2(t)dt,
\end{equation}
where $\dddot{l}$ is lateral jerk, and $\dddot{s}$ is longitudinal jerk.

The consistency between automation and driver intention is reflected in two aspects: the alignment of the driver's desired lane changes with the automation's planned trajectory and the alignment of the driver's desired velocity with the automation's planned velocity. The cost of consistency is computed using the following formulation:
\begin{equation}
\begin{aligned}C_\mathrm{consistency}(t)&=\int \left[l(t)-l_{target}(t)\right]^{2}dt\\&+\int \left[\dot{s}(t)-\dot{s}_{target}(t)\right]^{2}dt\end{aligned},
\end{equation}
where $l_{target}$ is  the lateral offset
of the lane centerline desired by the human driver from the reference line, and $\dot s_{target}$ is the target longitudinal velocity.

The solutions with slow convergence need to be penalized in the cost function to improve efficiency. The cost of efficiency is defined as:
\begin{equation}
    C_\mathrm{efficiency}(t)=T,
\end{equation}
where $T$ is the convergence time.

Taking all three factors into account, the final cost function is defined as:
\begin{equation}
\label{eq15}   C=w_{c1}C_\mathrm{comfort}+w_{c2}C_\mathrm{consistency}+w_{c3}C_\mathrm{efficiency},
\end{equation}
where $w_{c1}$, $w_{c2}$, and $w_{c3}$ denote the weights of comfort, consistency, and efficiency, respectively.

\subsubsection{Overall Algorithm}
To clarify, we present the complete trajectory planning algorithm in Algorithm~\ref{alg1}. First, sampling is performed in the lateral direction based on the road width. For each lateral position $d_i$, the target lateral offset $l(t_i)=d_i$, lateral velocity $\dot l(t_i)$, and lateral acceleration $\ddot l(t_i)$ are computed at each prediction time $t_i$. A lateral trajectory is then generated using a quintic polynomial, considering the initial and target states. Next, based on the target velocity sampling $v_i$, a longitudinal trajectory is generated using a quartic polynomial using the initial state $[s(t_0), \dot s(t_0), \ddot s(t_0)]$, the longitudinal target velocity $\dot s(t_i)=v_i$, and longitudinal acceleration $\ddot s(t_i)$. The cost of each trajectory is computed according to factors such as comfort, consistency, and efficiency, and both the trajectory and its cost are added to the set. Finally, all trajectories are checked, and the optimal trajectory with the minimum cost is selected from the remaining valid trajectories.

\subsection{RL-Based Authority Allocation}
\subsubsection{Markov Decision Process}
The Markov Decision Process (MDP) framework is used to model reinforcement learning problems. Specifically, an agent makes a decision and takes an action based on the current environment state. The environment then provides feedback in the form of a reward and transitions to next state according to the state transition probability. An MDP is described by a tuple $\left \langle S, A, P, r, \gamma \right \rangle$, where $S$ is the state space, $A$ is the action space, $P: S\times S\times A\rightarrow \mathbb{R}$ represents the state transition probability, $r: S\times A\rightarrow \mathbb{R}$ represents reward function for each transition, and $\gamma$ is the discount factor. The goal of an MDP is to maximize expected return through an optimal policy $\pi^*(a_t |s_t)$, which is written as:
\begin{equation}
\label{eq17}
    \pi^*=\underset{\pi}{\operatorname*{\operatorname*{argmax}}}\mathbb{E}_{(s_t,a_t)\sim\pi}\left[\sum_{t=0}^\infty\gamma^tr(s_t,a_t)\right],
\end{equation}
where $s_t\in S$ represents environment state at time $t$, $a_t\in A$ represents action at time $t$.

\subsubsection{Input State and Output Action}
The front wheel steering angles provided by the driver and the automation implicitly reflect their respective driving characteristics. Additionally, both driver state and vehicle state influence the vehicle actual trajectory. Therefore, the designed input state vector is defined as: 
\begin{equation}
    s_t=[\delta_a,\delta_h,s_h,s_v],
\end{equation}
where $\delta_a$ is the automation front wheel steering angle, $\delta_h$ is the driver front wheel steering angle, $s_h$ represents the quantified driver state, and $s_v$ represents the vehicle state. Specifically, the vehicle state $s_v$ includes lateral acceleration $a_{y}$, lateral offset $e_{d}$, and yaw angle error $e_{yaw}$.

The output action vector ${a_t} = [\lambda]$ is the automation control authority which is a continuous value in the range $\lambda \in [0,1]$.

\subsubsection{Reward Function}
The key to achieving the RL-based authority allocation strategy lies in a well-designed reward function that simultaneously guides the agent to maximize rewards and reach the desired goal. The goal is to ensure tracking performance, comfort and no collision while minimizing human-machine conflict. 

The tracking reward is based on the vehicle lateral offset $e_{d}$ and yaw angle error $e_{yaw}$. A smaller absolute value of $e_{d}$ and $e_{yaw}$ signifies better tracking precision. Thus, the tracking reward $r_\mathrm{tracking}$ is formulated as:
\begin{equation}
\label{eq18}
    r_\mathrm{tracking}=-(|e_{d}|+|e_{yaw}|).
\end{equation}

The comfort reward is composed of lateral acceleration $a_{y}$, lateral jerk $\dot a_{y}$, and the rate of change of the automation authority $\dot \lambda$. Smaller absolute values of these terms indicate improved comfort. The comfort reward $ r_\mathrm{comfort}$ is defined as: 
\begin{equation}
\label{eq19}
    r_\mathrm{comfort}=-(|a_{y}|+|\dot{a}_{y}|+|\dot{\lambda}|).
\end{equation}

If a collision occurs, the collision reward $r_\mathrm{collision}$ imposes a significant penalty:
\begin{equation}
\label{eq20}
    r_\mathrm{collision~}=-200,~~\mathrm{if~~collision}.
\end{equation}

The human-machine conflict is a critical factor to consider in the reward function. On one hand, the vehicle final steering angle should align as closely as possible with the driver steering characteristics. On the other hand, control authority should adapt to the driver state, with the automation being assigned higher control authority to enhance driving safety when the driver is in an abnormal state, such as distraction. Therefore, the human-machine conflict reward $r_\mathrm{conflict}$ is defined by the discrepancy between the vehicle final front wheel steering angle $\delta$ and the driver front wheel steering angle $\delta_h$, as well as the discrepancy between the quantified driver state $s_h$ and the automation control authority $\lambda$. The smaller the absolute values of these discrepancies, the lower the human-machine conflict. The human-machine conflict reward $ r_\mathrm{conflict}$ thus can be expressed as
\begin{equation}
\label{eq21}
    r_\mathrm{conflict}=-(|\delta-\delta_h|+|s_h-\lambda|).
\end{equation}

Combining all the individual rewards, the overall reward function is defined as:
\begin{equation}
\label{eq22}
    r=w_{r1}r_\mathrm{tracking}+w_{r2}r_\mathrm{comfort}+w_{r3}r_\mathrm{collision}+w_{r4}r_\mathrm{conflict},
\end{equation}
where $w_{r1}$, $w_{r2}$, $w_{r3}$, and $w_{r4}$ are the weights assigned to each reward, reflecting their relative importance in the overall objective.

\subsubsection{PPO-based Optimization Algorithm}
One approach to solving the optimal policy in Equation \eqref{eq17} is the policy gradient method, which directly models the policy function and optimizes it by maximizing the expected return using gradient ascent. Proximal Policy Optimization (PPO) \cite{ppo_schulman2017proximal}, as a specific policy gradient method, is widely recognized for its efficiency and stability in policy updates. Furthermore, PPO is well-suited for problems involving continuous action spaces and performs well in complex environments. 

The framework of PPO is built on the Actor-Critic architecture, where the actor network updates the policy and the critic network provides more accurate state value estimations. The objective function of actor network is defined as follows:
\begin{equation}\label{eq25}
\begin{aligned}
    J(\theta)=\mathbb{E}_{(s_t,a_t)\sim\pi_{\theta_{old}}}[\min(&\rho_t(\theta)A_t^{\theta_{old}}, \\
    &\mathrm{clip}(\rho_t(\theta),1-\varepsilon,1+\varepsilon)A_t^{\theta_{old}})],
\end{aligned}
\end{equation}
where $\rho_{t}(\theta)=\frac{\pi_{\theta}(a_{t}|s_{t})}{\pi_{\theta_{old}}(a_{t}|s_{t})}$ represents the ratio of the new policy $\pi_\theta(a_t|s_t)$ and the old policy $\pi_{\theta_{old}}(a_t|s_t)$ at time $t$, $A_t^{\theta_{old}}$ represents an estimator of the advantage function at time $t$, and $\mathrm{clip}(\cdot)$ is a clip function that restricts $\rho_t(\theta)$ to the interval $[1+\varepsilon,1-\varepsilon]$.

The critic network is optimized by minimizing the mean squared error between the estimated state value and the target state value. The corresponding loss function is defined as:
\begin{equation}\label{eq26}
    L(\varphi)=\mathbb{E}[(V(s_t)-V^{target})^2],
\end{equation}
where $V(s_t)$ is the value estimate of state $s_t$ calculated by the critic network, with the network parameters denoted by $\varphi$, and $V^{target}$ denotes the target state value. 

The training process of the PPO-based authority allocation strategy is summarized in Algorithm~\ref{alg2}. The agent interacts with the environment using the current policy, collects a certain number of transitions into the experience replay buffer, and then advantage estimates are computed using Generalized Advantage Estimation (GAE). Within the specified number of iterations, the ratio between the old and current policies is calculated. The objective function of the actor network and the loss function of the critic network are computed, and the parameters of both networks are updated using gradient ascent and gradient descent, respectively. Finally, the parameters of the new actor network are assigned to the old actor network, which helps improve the agent's policy performance by allowing the new and old actor networks to alternately collect data and update the networks.

\begin{algorithm}[t!]
\caption{PPO-Based Authority Allocation}\label{alg2}
\KwIn{driver front wheel steering angle $\delta_h$, automation front wheel steering $\delta_a$, driver state $s_h$, and vehicle state $s_v$}
\KwOut{the automation control authority $\lambda$}
Initialize: actor network $\pi_\theta$, critic network $V_\varphi$, experience replay buffer $B$ and environment\;
\For {episode $\leftarrow$ 1 \KwTo M}{
    Receive initial state\;
    \For {t $\leftarrow$ 1 \KwTo T}{
        Select action $a_t$ according to the current policy $\pi_{\theta_{old}}$\;
        Execute action $a_t$ and observe reward $r_t$ and new state $s_{t+1}$\;
        Store transition tuple $\{s_t, a_t, r_t, s_{t+1}\}$ into $B$\;
        Compute advantage estimates $A_t^{\theta_{old}}$\;

    }
    \For {epoch $\leftarrow$ 1 \KwTo K}{
        Compute the ratio $\rho_{t}(\theta)$\;
        Update  $\theta$ by maximizing $J(\theta)$ defined in Equation (\ref{eq25}) using gradient ascent\;
        Update  $\varphi$ by minimizing $L(\varphi)$ defined in Equation (\ref{eq26}) using gradient descent.
    }
    $\theta_{old} \leftarrow \theta$

}
\end{algorithm}

\section{Experiments}

\subsection{Experimental Setup}

\subsubsection{Structured Scenarios}
\begin{figure}[!t]
\centering
\includegraphics[width=1.0\columnwidth]{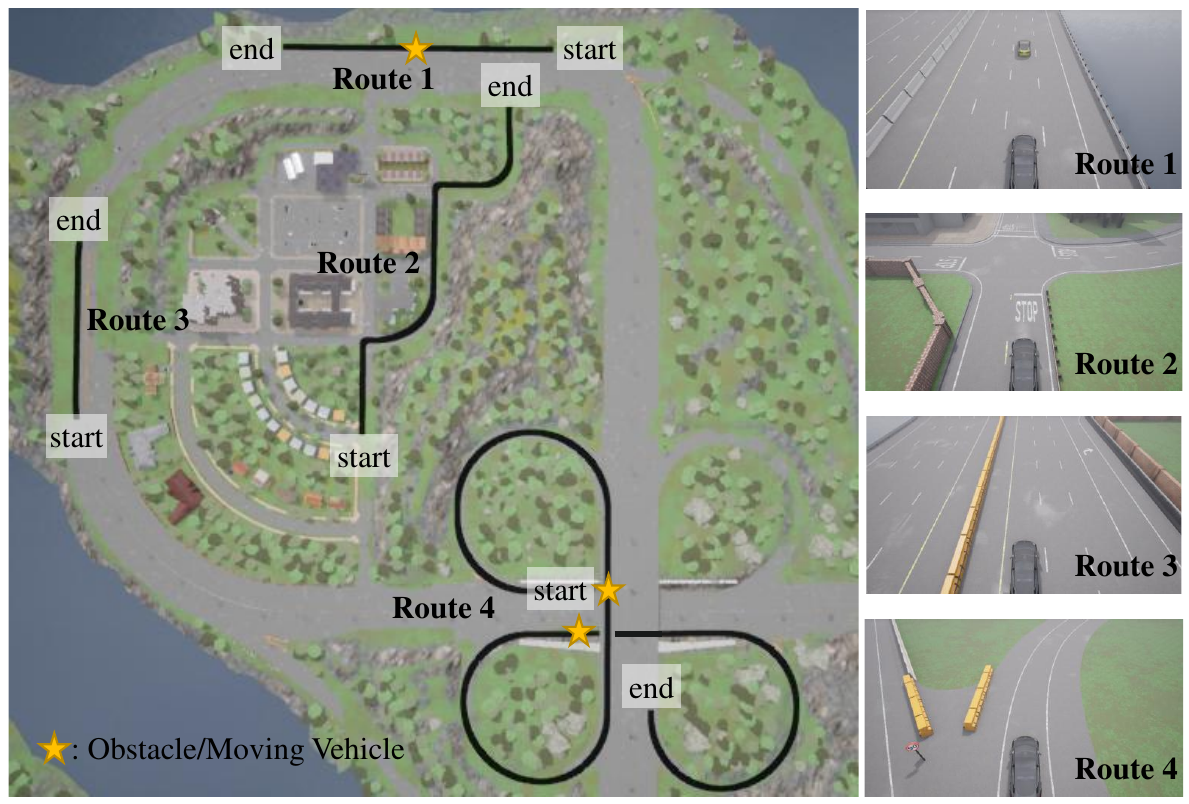}
\caption{Scenarios and routes in the CARLA driving simulator.}
\label{fig4}
\vskip -0.1in
\end{figure}

CARLA \cite{56dosovitskiy2017carla}, one of the mainstream simulators, can simulate diverse environmental conditions, construct various city layouts, and integrate elements such as buildings, pedestrians, road signs, and more. It also supports flexible sensor configurations and provides detailed data, including GPS coordinates, velocity, acceleration, and collision information. In this study, we use the CARLA simulator to construct structured road environments. As illustrated in Fig.~\ref{fig4}, the Town04 map reflects typical structured scenarios and features complex road topologies, such as one-way streets, viaducts, and beltways.

\begin{itemize}
    \item \textbf{Route 1}: A multi-lane scenario, with the vehicle in the middle lane. Both adjacent lanes are drivable, with an obstacle or moving vehicle present on the road. 
    
    \item \textbf{Route 2}: A combination of straight lanes, small-curvature curved lanes, and large-curvature curved lanes.
    
    \item \textbf{Route 3}: A multi-lane scenario with the vehicle in the leftmost lane, while the adjacent right lanes remain drivable. 
    
    \item \textbf{Route 4}: A combination of straight lanes and looped lanes, with moving vehicles on the road. 
    
\end{itemize}

\subsubsection{Evaluation Metrics}
To evaluate the interaction between the automation and driver, as well as the performance of the human-vehicle cooperative driving, we use six metrics that assess our proposed approach from multiple perspectives.

\textbf{Safety}: Trajectory tracking ability is vital to ensuring that the vehicle remains within safe boundaries and avoids collisions, which is often used as a measure of safety \cite{23_2022adaptive}. The safety metric is defined  as the average sum of the absolute values of the lateral offset $e_d$ and the yaw angle error $e_{yaw}$ over the time interval [0, T]:
\begin{equation}
    J=\frac{\int_0^T\lvert e_d\rvert dt+\int_0^T\lvert e_{yaw}\rvert dt}{T}.
\end{equation}

\textbf{Stability}: Vehicle stability is evaluated using lateral acceleration $a_y$ and lateral velocity $v_y$, as proposed in \cite{9fang2023human}. The stability metric is defined as:
\begin{equation}
    J=\frac{\int_0^T\lvert a_y\rvert dt+\int_0^T\lvert v_y\rvert dt}{T}.
\end{equation}

\textbf{Comfort}: Jerk is a key factor in determining comfort and is thus used as a metric to evaluate vehicle comfort:
\begin{equation}
    J=\frac{\int_0^T|jerk|dt}{T}.
\end{equation}

\textbf{Driver Physical Workload (DPW)}: The driver physical energy is expended when turning the steering wheel\cite{42wang2016human}. A driver physical energy consumption is influenced by the steering wheel angle and the frequency of its changes. Larger steering wheel angles and higher rates of angle change result in increased energy consumption. To quantify driver physical workload, the  front wheel steering angle $\delta_h$ and its rate of change $\dot\delta_h$ are used as the metric \cite{43bolia2014driver, 44powell2011relationships, 45deng2022shared}:
\begin{equation}
    J=\frac{\int_0^T\lvert\delta_h\rvert dt+\int_0^T\lvert\dot{\delta_h}\rvert dt}{T}.
\end{equation}

\textbf{Driver Cognitive Workload (DCW)}: Cognitive workload refers to the mental resources required for focused attention, situational awareness, decision-making, and execution. It is a key metric for evaluating human-machine systems \cite{46jin2025impact,47li2009synthetic, 48yan2024human}. Driver cognitive workload is indirectly assessed through performance in the non-driving-related task (NDRT), defined as: 
\begin{equation}
\label{eq29}
    J=\frac{1}{N}\sum t_{act},
\end{equation}
where $N$ is the total number of NDRTs, and $t_{act}$ is the driver reaction time in each NDRT. In the human-in-the-loop experiment, the NDRT requires the driver to press the corresponding button displayed and highlighted on the indicator, as illustrated in Fig.~\ref{fig9}.

\textbf{Human-Machine Conflict (HMC)}: The degree of divergence between the driver front wheel steering angle $\delta_h$ and the actual front wheel steering angle $\delta$ serves as a measure of human-machine conflict \cite{49zhang2021driving}. It can be quantified as:
\begin{equation}
    J=\frac{\int_0^T|\delta_h-\delta|dt}{T}.
\end{equation}

\subsection{Implementation Details}
\subsubsection{Driver Model and Vehicle Controller}
The driver model simulates the human driver steering behavior and interacts with the automation during the training of the reinforcement learning agent,  characterizing driver operations. In this paper, the two-point preview model \cite{42wang2016human, 50sentouh2018driver} is adopted as the driver model, which simulates the driver adjustment of the steering angle based on near and far preview points. 
Additionally, the driver neural reaction delay time varies across different states, such as concentrated, normal, and distracted. For a driver in a normal state, the neural reaction delay time is typically around 0.3 seconds \cite{52guo2008}. 
We simulate the driver in different states by changing the reaction delay time based on the two-point preview model. For a concentrated driver, the reaction delay time is set to 0.2 seconds, while for a distracted driver, it is set to 0.5 seconds. These reaction times are integrated into the two-point preview model to simulate driver behavior under different driver states.

Model Predictive Control (MPC) is an effective control method widely applied to vehicle tracking and control, due to its ability to handle nonlinear systems and predict future states \cite{53falcone2007predictive,shen2021distributed,yin2022distributed}. In this study, the MPC controller is designed based on a two-degree-of-freedom vehicle dynamics model, which is an essential simplified representation widely used to study vehicle motion. 
After linearizing and discretizing the control model, the deviation at each sampling time is optimized through rolling horizon optimization, ensuring precise tracking of the desired trajectory.

\subsubsection{Reinforcement Learning Agent Selection}
To identify the optimal agent, we evaluated three efficient reinforcement learning agents: Soft Actor-Critic (SAC) \cite{sac_haarnoja2018soft}, Deep Deterministic Policy Gradient (DDPG) \cite{ddpg_lillicrap2015continuous}, and Proximal Policy Optimization (PPO) \cite{ppo_schulman2017proximal}. Among these, DDPG is a deterministic policy-based algorithm. 
The hyperparameters of the RL agent were selected through careful tuning and sensitivity analysis to ensure robust performance. The final hyperparameter settings are summarized in Table~\ref{table1}.
Rectified Linear Units (ReLU) are used in the activation layers, and the Adam optimizer is employed for network training.

\begin{table}[!t]
\caption{hyperparameter configuration}
\centering
\resizebox{\columnwidth}{!}{
\begin{tabular}{cccc}
\toprule
        Hyperparameter & PPO & SAC & DDPG \\
\midrule
        Replay Buffer Size & 100000 & 100000 & 100000 \\
        Batch Size & 128 & 128 & 128 \\
        Hidden Layer Dimension & 256 & 256 & 256 \\
        Discount Factor & 0.90 & 0.90 & 0.90 \\
        Actor Learning Rate & 0.0003 & 0.0003 & 0.0003 \\
        Critic Learning Rate & 0.0003 & 0.0003 & 0.003 \\
        Target Smoothing Factor & - & 0.005 & 0.005 \\
        Explore Noise & - & - & N $\sim$ (0.01,0.2) \\
        Lambda for GAE & 0.95 & - & - \\
        Clip Factor & 0.2 & - & - \\
        Entropy Coefficient & 0.01 & - & - \\
        Optimization Epoch & 10 & - & - \\
\bottomrule
\end{tabular}
}\label{table1}
\end{table}

\begin{figure}[!t]
\centering
\includegraphics[width=\linewidth]{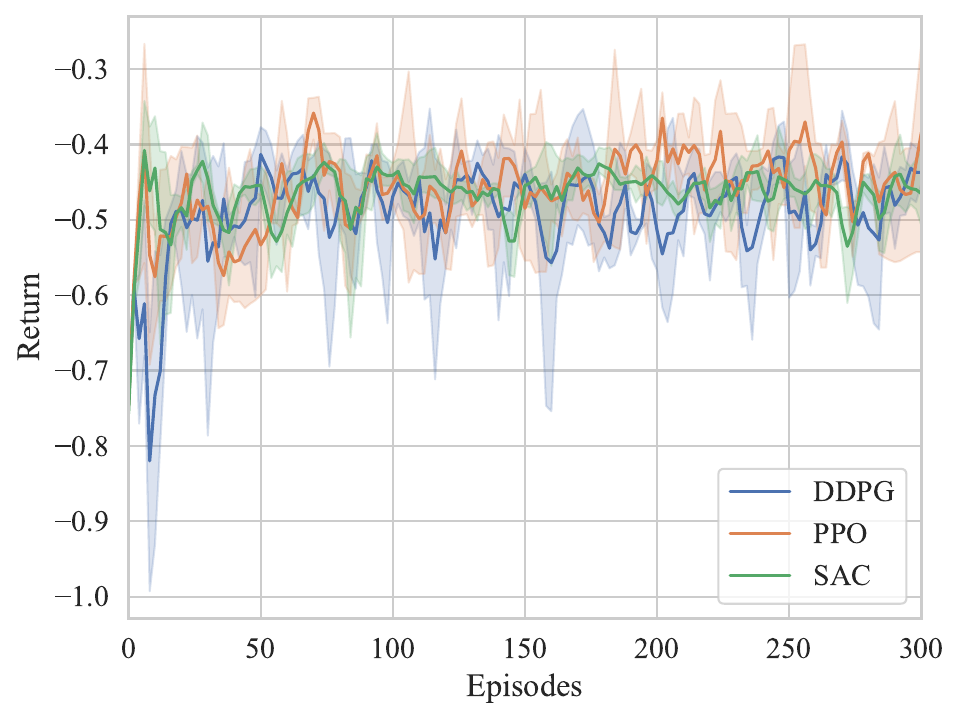}
\vskip -0.1in
\caption{Performance comparison curves for SAC, PPO, and DDPG.}
\label{fig5}
\vskip -0.1in
\end{figure}

Fig.~\ref{fig5} shows the total return during evaluation for PPO, SAC, and DDPG. Each agent is trained with different random seeds, with one evaluation conducted every two episodes. The solid curves represent the average return, and the shaded regions indicate the standard deviation. The results demonstrate that SAC and PPO exhibit faster convergence rates than DDPG, with PPO exhibiting a higher average return than both DDPG and SAC overall. Given that PPO obtains the optimal policy with limited training steps and yields the highest return, we select the PPO agent as the authority allocation strategy for subsequent experiments.

\subsubsection{Comparative Methods}
We compare our HOCD approach against three other approaches. These approaches and settings are outlined as follows:
\begin{itemize}
\item Manual Driving (MD): The vehicle is entirely controlled by the driver without any assistance from automation. In simulation experiments, the driver refers to the driver model, whereas in the human-in-the-loop experiments, the real human is involved. 
\item Fixed-Authority-Based Cooperative Driving (FACD): The driver and the automation cooperate with a predefined control authority. When the driver is concentrated, normal, or distracted, the automation control authority is set to 0.2, 0.5, or 0.8, respectively.
\item Driver-Characteristics-based Cooperative Driving (DCCD)~\cite{9fang2023human}: The authority allocation strategy is based on the exponential function as follows: 
\begin{equation}
    \lambda=\left\{\begin{array}{cc}\lambda_{min},&\lambda\leq\lambda_{min}\\e^{-(\mu_1DI)^{\mu_2}(\mu_3DA)^{\mu_4}},&\lambda\geq\lambda_{min}\end{array}\right.,
\end{equation}
where the minimum automation control authority $\lambda_{min}$ is set to 0.1, and the parameters are $\mu_1=2$, $\mu_2=3$, $\mu_3=1$, and $\mu_4=3$. DA represents driving ability and is calculated as:
\begin{equation}
    DA=\frac{1}{1+(\alpha_1e_d)^2+\left(\alpha_2e_{yaw}\right)^2},
\end{equation}
where $\alpha_1=0.75$ and $\alpha_2=0.22$ are weight coefficients for the lateral offset $e_d$ and yaw angle error $e_{yaw}$. DI denotes the driver involvement, which varies according to the driver state. In this paper, when the driver is concentrated, normal, or distracted, DI is set to 0.6, 0.45, or 0.3, respectively.
\end{itemize}

\subsection{Simulation Experiments}
\subsubsection{Trajectory Planning Based on Driver Intention}

\begin{figure}[!t]
\centering
\subfloat[Actual trajectories corresponding to different driver intentions.]{\includegraphics[width=1.0\columnwidth]{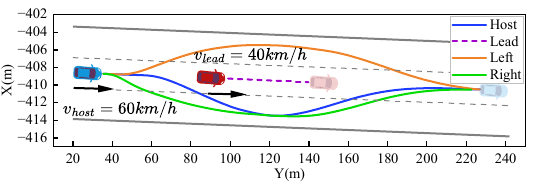}
\label{fig6-a}}
\hfil
\subfloat[Trajectory sampling based on the driver left-turn intention]{\includegraphics[width=1.0\columnwidth]{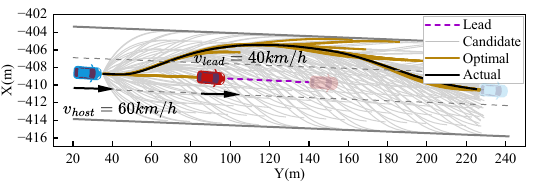}
\label{fig6-b}}
\caption{Visualization of intention-aware trajectory planning.}
\label{fig_sim}
\vskip -0.1in
\end{figure}


\begin{figure*}[!t]
\centering
\subfloat[Concentrated]
{\includegraphics[width=0.65\columnwidth]{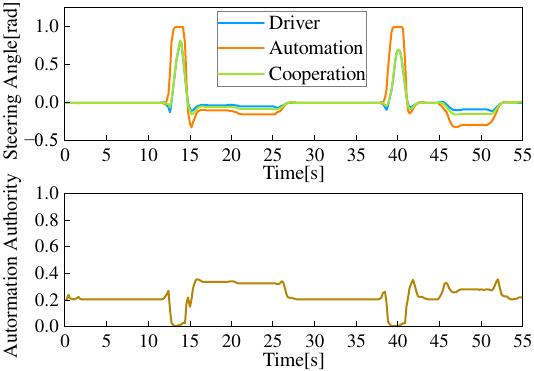}
\label{fig7-a}}
\hfil
\subfloat[Normal]{\includegraphics[width=0.65\columnwidth]{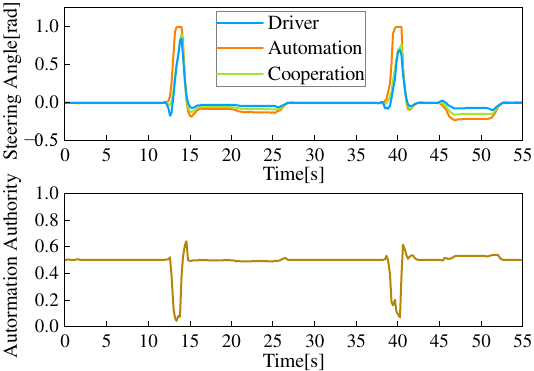}
\label{fig7-b}}
\hfil
\subfloat[Distracted]{\includegraphics[width=0.65\columnwidth]{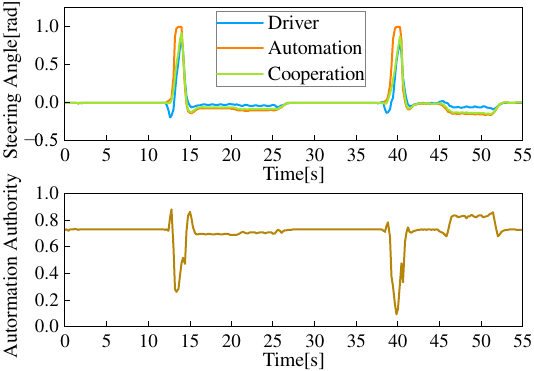}
\label{fig7-c}}
\caption{Visualization of steering angle and authority allocation under different driver states based on the driver model.}\label{fig7}
\vskip -0.1in
\end{figure*}

Fig.~\ref{fig6-a} shows the different trajectories of the host vehicle on Route 1 when encountering a slow vehicle ahead, influenced by varying driver intentions. The blue trajectory represents the path planned by the automation without considering the driver intention. In contrast, the green trajectory corresponds to the driver's intention to change lanes in advance, while the yellow trajectory reflects the intention to change lanes from the left.
Fig.~\ref{fig6-b} visualizes the sampling process for the driver's intention to change lanes from the left, showing combined lateral and longitudinal trajectory sets in the Cartesian frame. The gray curves represent all candidate trajectories, the brown curves indicate the optimal trajectory with the lowest cost, and the black trajectory corresponds to the actual trajectory. It can be concluded that the trajectory planning method we developed is capable of effectively generating an optimal trajectory that aligns with driver intention, ensuring seamless integration between driver input and the automation planning process. 

\subsubsection{Authority Allocation Based on Driver State}

Fig.~\ref{fig7} presents the results of PPO-based authority allocation and the variation in steering angle under different driver states on Route 2. Overall, the proposed strategy dynamically adjusts the level of automation control authority based on driver state.  As the driver's attention decreases, a greater level of control authority is allocated to the automation. As shown in Fig.~\ref{fig7-a}, when the driver is fully concentrated, the automation is assigned a lower control authority of approximately 0.2, allowing the driver to retain primary control. Fig.~\ref{fig7-b} shows that when the driver is in a normal state, the automation control authority fluctuates around 0.5. As illustrated in Fig.~\ref{fig7-c}, when the driver is distracted, the automation holds a dominant position, with its authority reaching around 0.8, which indicates that increased automation intervention is required to ensure overall safety.

Furthermore, the proposed strategy dynamically allocates authority during the entire driving process. At the bend in the road, the automation control authority first decreases and then increases. The decrease reflects the driver steering intention, which is prioritized to minimize human-machine conflicts. The subsequent increase indicates automation intervention, assisting the driver in steering more effectively.

It can be concluded that the proposed allocation strategy effectively balances the driver state, input, and automation intervention, fostering cooperation and minimizing potential conflicts between human and automation.

\subsubsection{Comparison of Human-Machine Conflicts}

\begin{figure}[!t]
\centering
\includegraphics[width=0.88\columnwidth]{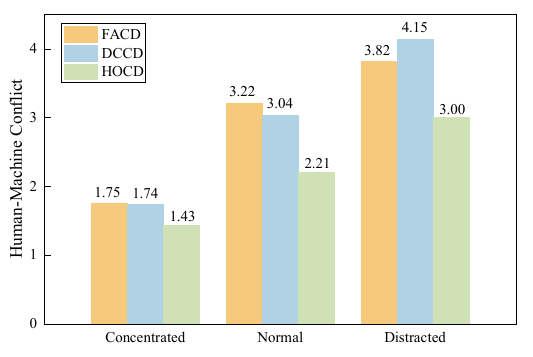}
\caption{Comparison of human-machine conflicts under different driver states.}
\label{fig8}
\vskip -0.13in
\end{figure}

Fig.~\ref{fig8} compares the human-machine conflict performance of three cooperative driving approaches: FACD, DCCD, and HOCD, across concentrated, normal, and distracted driver states.
In the concentrated state, FACD exhibits the highest human-machine conflict (1.75), followed closely by DCCD (1.74). In the normal state, the conflict values of FACD and DCCD are 3.22 and 3.04, respectively. In the distracted state, DCCD shows the highest conflict, with a value of 4.15, while FACD records a slightly lower conflict of 3.82. In contrast, the proposed HOCD approach consistently and significantly reduces human-machine conflict across all driver states, outperforming both FACD and DCCD. The superiority is likely due to the ability of HOCD to dynamically adjust the allocation strategy based on real-time feedback. In comparison, the function-based method (DCCD) and fixed-authority-based method (FACD) rely on preset rules, limiting flexibility and resulting in suboptimal performance.

\begin{table}[!t]
    \caption{Comparison of memory usage and inference time.}
    \centering
    \begin{tabular}{cccc}
    \toprule
        ~ & HOCD & DCCD & FACD  \\
    \midrule
        Memory (MB) & 0.1396 & 0.0054 & 0.0005  \\ 
        Inference Time (ms) & 1.1799 & 0.4854 & 0.0002 \\
    \bottomrule
    \end{tabular}
\label{table2}
\end{table}

\subsubsection{Comparison of Computational Efficiency}

We evaluate memory consumption and inference time among FACD, DCCD, and HOCD. As shown in Table~\ref{table2}, the memory consumption of HOCD is 0.1396 MB, and the average inference time per step is 1.1799 ms, whereas DCCD and FACD require significantly fewer computational resources. This is primarily because RL-based HOCD relies on deep neural networks for policy learning, which involve more parameters and higher computational load compared to the simpler function-based DCCD and fixed-authority-based FACD.



Despite the increased computational cost, the superior adaptability and performance of HOCD justify this overhead. Its learning-based design allows for effective modeling of complex scenarios and ensures robust operation in dynamic environments. Notably, the typical control cycle of autonomous driving systems is 10 ms, and the inference time of HOCD is 1.1799 ms, which falls well within this limit, demonstrating its practicality for real-time deployment.


\begin{figure}[!t]
\centering
\includegraphics[width=0.83\columnwidth]{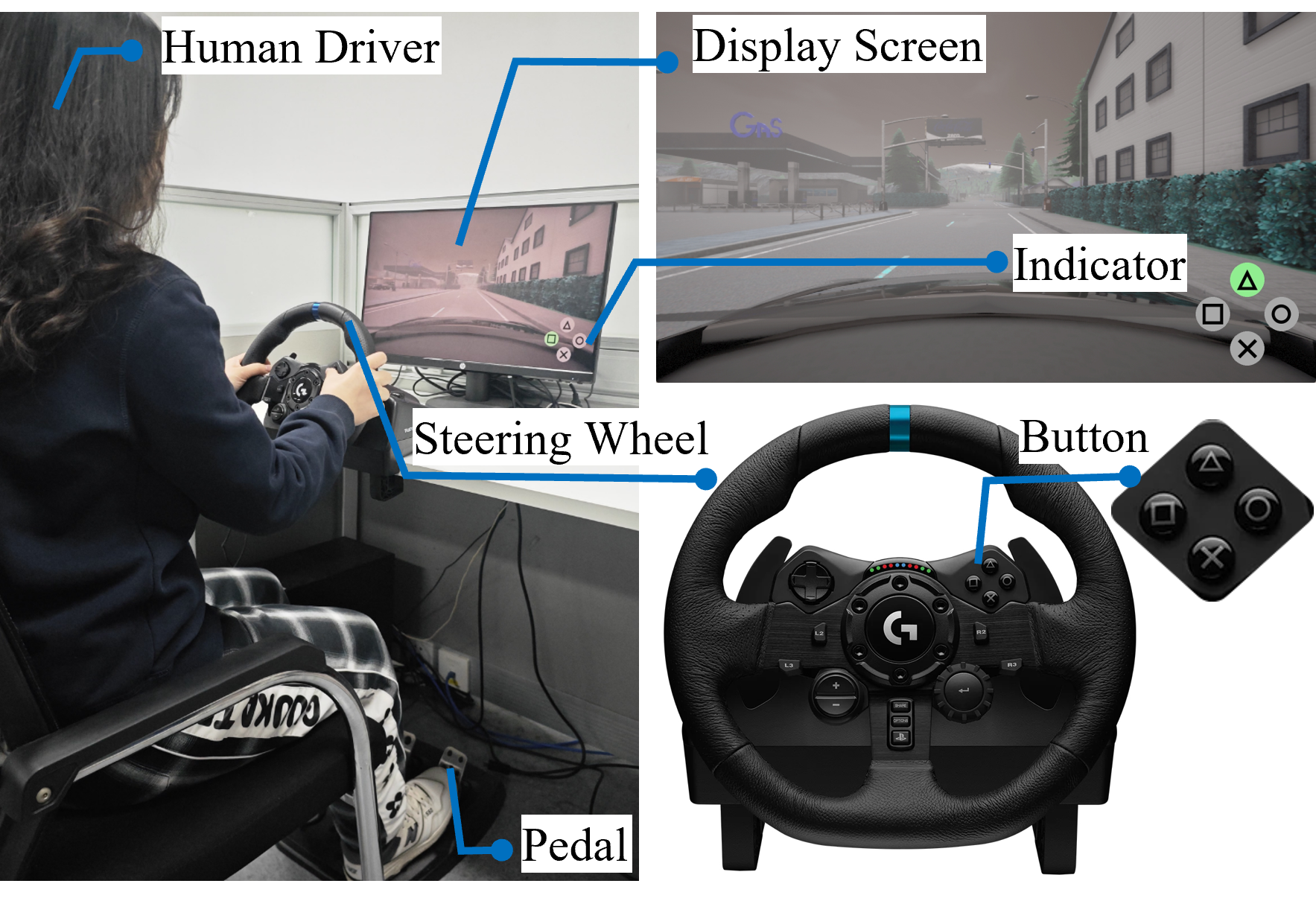}
\caption{Experimental platform for human-in-the-loop experiments.}
\label{fig9}
\vskip -0.1in
\end{figure}

\subsection{Human-in-the-Loop Experiments}
To better simulate and evaluate the interaction between the human driver and automation, we designed a human-in-the-loop driving simulator, as illustrated in Fig.~\ref{fig9}. The front screen displays the driving environment generated by CARLA, while the driver model is replaced by a real human driver, who provides steering wheel angle inputs using the Logitech G923 steering wheel.

\begin{figure*}[!t]
\centering
\subfloat[Concentrated]
{\includegraphics[width=0.65\columnwidth]{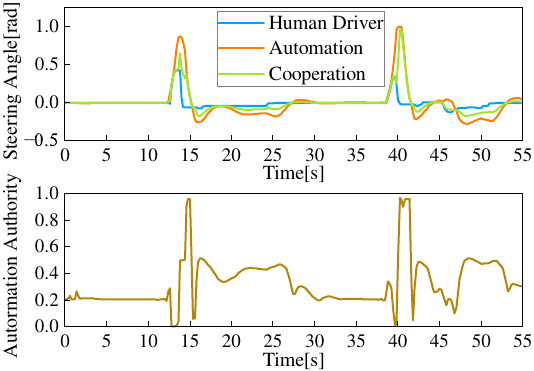}
\label{fig10-a}}
\hfil
\subfloat[Normal]{\includegraphics[width=0.65\columnwidth]{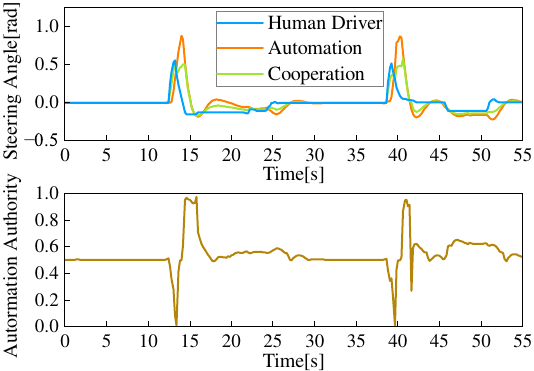}
\label{fig10-b}}
\hfil
\subfloat[Distracted]{\includegraphics[width=0.65\columnwidth]{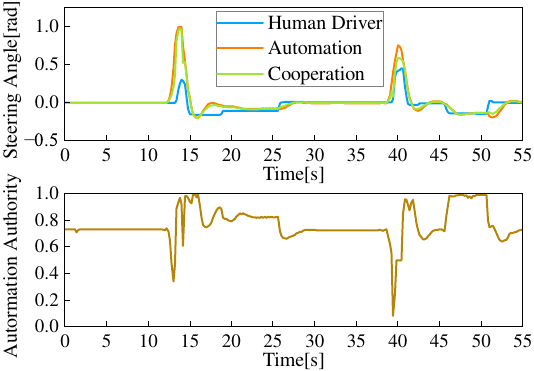}
\label{fig10-c}}
\caption{Visualization of steering angle and authority allocation under different human driver states.}\label{fig10}
\vskip -0.1in
\end{figure*}

\begin{figure*}[!t]
\centering
\subfloat[Scenario 1: lane departure suppression]{\includegraphics[width=1.0\columnwidth]{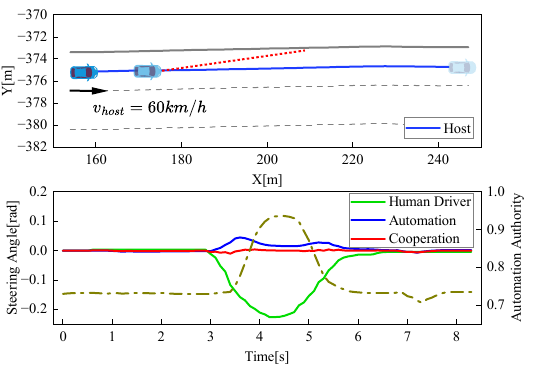}
\label{fig11-a}}
\hfil
\subfloat[Scenario 2: cooperative driving]{\includegraphics[width=1.0\columnwidth]{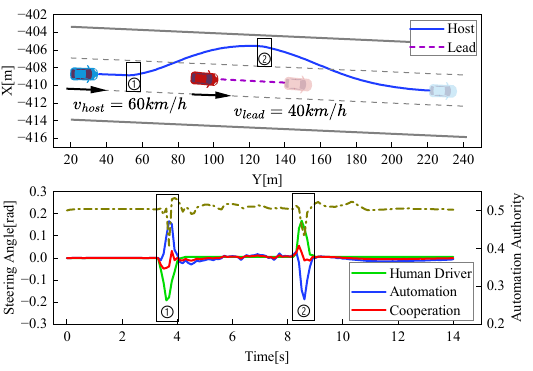}
\label{fig11-b}}
\hfil
\subfloat[Scenario 3: blind spot hazard avoidance]{\includegraphics[width=1.0\columnwidth]{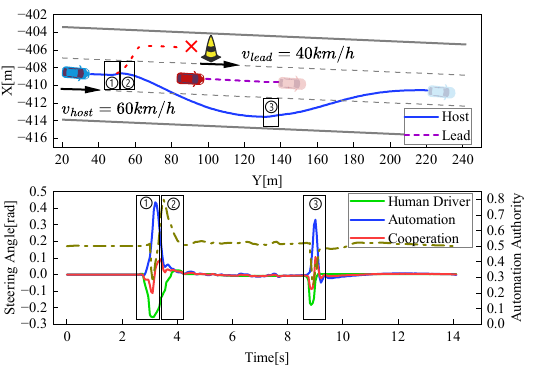}
\label{fig11-c}}
\hfil
\subfloat[Scenario 4: emergent driving scenarios]{\includegraphics[width=1.0\columnwidth]{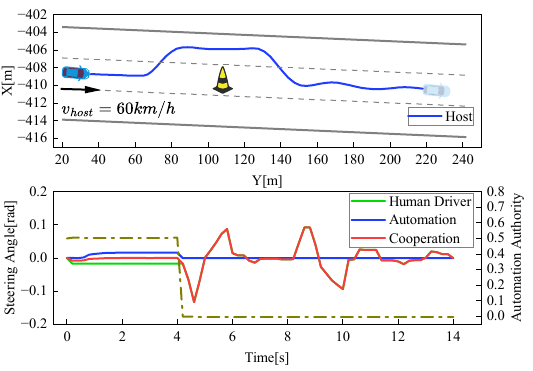}
\label{fig11-d}}
\hfil
\subfloat[Scenario 5: multi-vehicle conflict mitigation]{\includegraphics[width=1.0\columnwidth]{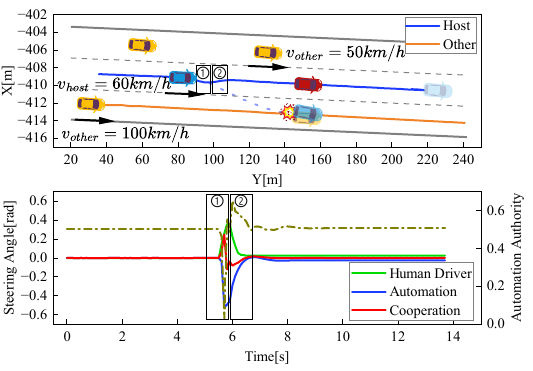}
\label{fig11-e}}

\caption{Visualization of trajectories, steering angles, and authority allocation in different scenarios. Each subfigure (a)–(e) includes an upper plot of the vehicle trajectory and a lower plot of the steering angle and authority. In the lower plots, the red, green, and blue lines denote the steering angles of the final cooperative output, the human driver, and the automation, respectively, while the dashed line indicates the automation control authority.}\label{fig11}
\vskip -0.1in
\end{figure*}

\subsubsection{Authority Allocation Validation Experiments} 
We first validate the robustness of the PPO-based authority allocation strategy with a real human driver in the loop. 

As illustrated in Fig.~\ref{fig10}, the authority allocation trends in the real human driving experiment are consistent with those obtained from the driver-model-based simulations in Fig.~\ref{fig7}. Specifically, when the driver's attention decreases, the automation is assigned higher control authority to ensure safety, whereas when the driver is concentrated, more authority is returned to the human driver.

In curved road segments, similar dynamic adjustment characteristics are also observed in the human-in-the-loop experiment. The automation authority decreases at the beginning of the curve to prioritize the driver’s steering intention and subsequently increases to assist the driver in completing the steering task. This behavior pattern also aligns with the driver-model-based results presented in Fig.~\ref{fig7}, further demonstrating that the proposed authority allocation strategy exhibits applicability and stability in real human-involved driving scenarios.

\subsubsection{Fixed Scenario Experiments} We further demonstrate the effectiveness of HOCD by carrying out experiments in five different scenarios.

\textbf{Scenario 1: lane departure suppression.}
In Route 3, the host vehicle travels at 60 km/h in the leftmost lane. Due to the driver  distraction, there is a risk of the vehicle departing from the lane. As illustrated in the top subplot of Fig.~\ref{fig11-a}, the red dashed line indicates the trajectory that the driver would follow due to an incorrect maneuver, while the blue line depicts the actual cooperative trajectory. As shown in the bottom subplot of Fig.~\ref{fig11-a}, when the driver begins to deviate from the lane, the absolute value of the driver steering angle gradually increases due to the incorrect maneuver. At the same time, the automation applies an opposite steering angle to counteract the driver input, leading to a sharp increase in the automation control authority to keep the vehicle within the lane and mitigate potential danger. Once the driver realizes the mistake and resumes normal driving, the automation opposing steering angle decreases, and its control authority gradually declines. Overall, the automation control authority exhibits a trend of initial decline followed by a subsequent increase during this process. The above results demonstrate that the proposed method can effectively mitigate the risk of lane departure.

\textbf{Scenario 2: cooperative driving.}
In Route 1, the host vehicle travels at 60 km/h and intends to change lanes to overtake a slower vehicle ahead traveling at 40 km/h, as shown in Fig.~\ref{fig11-b}. Overall, since the driver is in a normal state, the automation control authority remains at approximately 0.5, with small fluctuations due to the system adaptive adjustments to assist the driver in maintaining control. Specially, at point \ding{172}, the driver signals an intention to change lanes to the left, and the automation checks the safety of the driver intention. Once it is confirmed to be safe, the automation decreases its control authority and plans a trajectory in accordance with the driver intention. At point \ding{173}, after overtaking the slower vehicle, the driver signals a desire to return to the original lane.  The automation determines that this intention is safe, then reduces its control authority and plans a trajectory consistent with driver intention. 
These results demonstrate that when conditions are deemed safe, the automation adapts to driver intention by reducing its control authority and offers subtle adjustments to support cooperative driving. Therefore, the proposed method effectively respects and accommodates driver intention during cooperative driving.

\textbf{Scenario 3: blind spot hazard avoidance.}
In Route 1, the host vehicle travels at 60km/h, while a slower vehicle ahead in the current lane moves at 40 km/h. The driver intends to overtake from the left lane but fails to notice a hazard ahead in that lane. As depicted in Fig.~\ref{fig11-c}, the red dashed line illustrates the trajectory that the driver plans to follow when attempting to overtake. At point \ding{172}, the driver, unaware of the hazard in the left lane, signals the lane change intention and changes lanes by turning the steering wheel. The automation initially aims to align with driver intention, resulting in a decrease in its control authority. However, at point \ding{173}, the automation detects a potential danger associated with overtaking to the left. To mitigate this risk, the automation plans an alternate trajectory to change to the right lane, sharply increasing its control authority to avoid the hazard. At point \ding{174},  after overtaking the slower vehicle, the driver signals an intention to return to the original lane. The automation verifies the safety of this maneuver and reduces its control authority to respect driver intention. During the remaining time, the automation provides assistance by making slight adjustments, with control authority fluctuating around 0.5. These results demonstrate that the proposed approach effectively helps the driver avoid potential hazards in the blind-spot scenario.

\textbf{Scenario 4: emergent driving scenarios.}
In Route 1, the host vehicle travels at 60 km/h and encounters an obstacle in its current lane, as illustrated in the top subplot of Fig.~\ref{fig11-d}. Approximately 40 meters before reaching the obstacle, the automation suddenly becomes unreliable (e.g., due to sensor or planning module failure). Under such circumstances, the system immediately alerts the driver and subsequently transfers full control authority. As shown in the bottom subplot of Fig.~\ref{fig11-d}, the automation control authority decreases from 0.5 to 0, enabling the driver to fully take over the vehicle and steer to avoid potential danger.
Additionally, when the automation operates reliably, the human-vehicle cooperative driving approach remains active. In this case, the driver, assisted by the automation, maintains a smoother trajectory and achieves better control performance. However, once the system becomes unreliable and the driver takes over full control, the trajectory becomes slightly unstable, and control performance degrades. This observation indirectly demonstrates the effectiveness of the human-vehicle cooperative driving approach in assisting the driver at the control level.

\textbf{Scenario 5: multi-vehicle conflict mitigation.}
In Route 1, the host vehicle attempts to overtake a slower vehicle ahead. The front-left lane is blocked by a vehicle traveling at 50 km/h, while the rear-right lane is occupied by a fast-approaching vehicle traveling at 100 km/h. Consequently, the host vehicle faces simultaneous conflicts in both lateral directions, as shown in Fig.~\ref{fig11-e}. However, the driver fails to recognize the unsafe gap in the right lane and initiates a rightward lane change, which would lead to a potential collision risk. At point \ding{172}, the automation initially reduces its control authority to align with the driver intention. At point \ding{173}, it then rapidly increases its authority to keep the vehicle on a straight trajectory, thereby preventing the potential conflict. Therefore, the proposed method effectively mitigates multi-vehicle conflicts through adaptive control authority allocation.

\begin{table}[!t]
    \caption{Natural situation of participants.}
    \centering
    \begin{threeparttable}
        \resizebox{\columnwidth}{!}{
        \begin{tabular}{cccccccc|c}
        \toprule
            Driver No. & 1 & 2 & 3 & 4 & 5 & 6 &7 & Avg. \\
        \midrule
            Age & 26 & 25 & 27 & 22  & 22 & 22 & 24 & 24 \\
            Sex & F & M & F & M & F & M & M & -\\
            Driving Experience & 5 & 6 & 1 & 3 & 4 & 3 & 3 & 3.57\\
        \bottomrule
        \end{tabular}\label{table3}
        }
        \begin{tablenotes} 
    		\item F represents Female, and M represents Male. 
         \end{tablenotes} 
    \end{threeparttable} 
    \vskip -0.1in
\end{table}

\subsubsection{User Study}

We recruited seven participants in a normal state to engage in cooperative driving with the automation on Route 4. The participants included three females and four males, with an average age of 24 years and an average driving experience of 3.57 years, as illustrated in Table~\ref{table3}. Prior to the experiment, participants were briefed on the following details:
\begin{itemize}
\item The driver is only responsible for lateral control using the steering wheel, while longitudinal velocity is automatically managed by a PID controller. 
\item The driver should strive to maintain the vehicle along the lane centerline and avoid crossing the lane boundaries.
\item The driver is required to change lanes to avoid danger when an obstacle or a slow vehicle appears ahead.
\item The driver is required to press the corresponding button on the steering wheel as quickly as possible when the indicator is highlighted on the display screen, as shown in Fig.~\ref{fig9}.
\end{itemize}

To ensure a fair comparison, participants are unaware of the specific driving approach being tested. The approaches, including manual driving (MD), fixed-authority-based cooperative driving (FACD), driver-characteristics-based cooperative driving (DCCD), and human-oriented cooperative driving (HOCD), are randomly assigned in each experiment. In addition, the indicator on the display screen is also randomly highlighted. Following the completion of each driving approach test, the driver is required to complete a questionnaire to assess their subjective perception. As illustrated in Table~\ref{table4}, the questionnaire includes three items, each rated on a 5-point Likert scale (0–4). 

\begin{table}[!t]
    \caption{questionnaire.}
    \centering
    \resizebox{\columnwidth}{!}{
    \begin{tabular}{cll}
    \toprule
        No. & Evaluation & Score Range \\
    \midrule
        1 & Level of system assistance & \makecell[l]{0 - Fully manual, \\ 4 - Fully automated} \\
        2 & Trust in the system during the NDRT & \makecell[l]{0 - Very untrustworthy,\\ 4 - Very trustworthy} \\
        3 & Overall satisfaction & \makecell[l]{0 - Very dissatisfied, \\4 - Very satisfied} \\
    \bottomrule
    \end{tabular}\label{table4}
    }
    \vskip -0.1in
\end{table}

\begin{table*}[!t]
\caption{Comparison of manual driving (MD) and human-oriented cooperative driving(HOCD)}
\centering
\resizebox{\textwidth}{!}{
\begin{tabular}{ccccccccccccc}
\toprule
        ~ & \multicolumn{2}{c}{Driver1} & \multirow{2}*{\makecell{Relative \\ Improvement}} & \multicolumn{2}{c}{Driver2} & \multirow{2}*{\makecell{Relative \\ Improvement}} & \multicolumn{2}{c}{Driver3} & \multirow{2}*{\makecell{Relative \\ Improvement}} &
        \multicolumn{2}{c}{Driver4} & \multirow{2}*{\makecell{Relative \\ Improvement}} \\
        ~ & MD & HOCD & ~ & MD & HOCD & ~ & MD & HOCD & ~& MD & HOCD & ~ \\
\midrule
        Safety & 19.484  & \textbf{9.852}  & 0.978  & 20.858  & \textbf{9.939}  & 1.099  & 25.801  & \textbf{9.600}  & 1.688  & 18.316  & \textbf{9.827}  & 0.864   \\
        Stability & 19.626  & \textbf{17.631}  & 0.113  & 20.371  & \textbf{17.677}  & 0.152  & 26.256  & \textbf{17.487}  & 0.501  & 19.216  & \textbf{17.521}  & 0.097   \\
         Comfort & 12.576  & \textbf{11.150}  & 0.128  & 12.717  & \textbf{11.195}  & 0.136  & 16.683  & \textbf{11.372}  & 0.467  & 12.166  & \textbf{11.232}  & 0.083   \\
        DPW & 8.417  & \textbf{7.701} & 0.093  & 8.654  & \textbf{8.154}  & 0.061  & 10.599  & \textbf{5.678}  & 0.867  & 8.346  & \textbf{7.182}  & 0.162   \\ 
        DCW & 0.711  & \textbf{0.638}  & 0.114  & 0.715  & \textbf{0.577}  & 0.239  & 0.821  & \textbf{0.741}  & 0.108  & 0.897  & \textbf{0.689}  & 0.302   \\ 
        
\toprule
        ~ & \multicolumn{2}{c}{Driver5} & \multirow{2}*{\makecell{Relative \\ Improvement}} & \multicolumn{2}{c}{Driver6} & \multirow{2}*{\makecell{Relative \\ Improvement}} & \multicolumn{2}{c}{Driver7} & \multirow{2}*{\makecell{Relative \\ Improvement}} &
        \multicolumn{2}{c}{Average} & \multirow{2}*{\makecell{Relative \\ Improvement}} \\
        ~ & MD & HOCD & ~ & MD & HOCD & ~ & MD & HOCD & ~& MD & HOCD & ~ \\
\midrule
        Safety & 23.476  & \textbf{9.838}  & 1.386  & 19.629  & \textbf{9.953}  & 0.972  & 30.206  & \textbf{9.512}  & 2.176  & 22.539  & \textbf{9.789}  & 1.303   \\ 
        Stability & 33.393  & \textbf{17.707}  & 0.886  & 20.347  & \textbf{17.138}  & 0.187  & 20.692  & \textbf{17.325}  & 0.194  & 22.843  & \textbf{17.498}  & 0.305   \\
        Comfort & 17.065  & \textbf{12.060}  & 0.415  & 12.349  & \textbf{11.877}  & 0.040  & 12.612  & \textbf{11.291}  & 0.117  & 13.738  & \textbf{11.454}  & 0.199   \\
        DPW & 12.730  & \textbf{6.843}  & 0.860  & 8.497  & \textbf{2.933}  & 1.897  & 8.772  & \textbf{4.454}  & 0.969  & 9.431  & \textbf{6.135}  & 0.537   \\
        DCW & 0.698  & \textbf{0.609}  & 0.146  & 0.931  & \textbf{0.695}  & 0.340  & 0.730  & \textbf{0.685}  & 0.066  & 0.786  & \textbf{0.662} & 0.188   \\

\bottomrule
\end{tabular}
}\label{table5}
\vskip -0.1in
\end{table*}

\begin{table*}[!t]
\caption{Comparison of different cooperative driving approaches}
\centering
\begin{tabular}{cccc|ccc|ccc|ccc}
\toprule
        ~ & \multicolumn{3}{c}{Driver1} & \multicolumn{3}{c}{Driver2} & \multicolumn{3}{c}{Driver3} & \multicolumn{3}{c}{Driver4}\\
        ~ & HOCD & DCCD & FACD & HOCD & DCCD & FACD & HOCD & DCCD & FACD & HOCD & DCCD & FACD \\
\midrule
        Safety & \textbf{9.852}  & 10.358  & 10.439  & \textbf{9.939}  & 11.831  & 12.809  & \textbf{9.600}  & 12.021  & 12.565  & \textbf{9.827}  & 10.099  & 10.041   \\
        Stability & 17.631  & \textbf{17.517}  & 17.700  & \textbf{17.677}  & 18.484  & 18.465  & \textbf{17.487}  & 18.733  & 18.567  & \textbf{17.521}  & 18.163  & 17.659   \\
        Comfort & 11.150  & \textbf{11.080}  & 11.090  & \textbf{11.195}  & 14.067  & 14.028  & \textbf{11.372}  & 11.442  & 11.410  & 11.232  & 14.191  & \textbf{11.057}   \\
        DPW & \textbf{7.701}  & 11.517  & 12.394  & \textbf{8.154}  & 21.551  & 30.626  & \textbf{5.678}  & 24.280  & 29.834  & \textbf{7.182}  & 7.319  & 9.258   \\
        DCW & \textbf{0.638}  & 0.687  & 0.679  & \textbf{0.577}  & 0.713  & 0.608  & \textbf{0.741}  & 0.894  & 0.852  & \textbf{0.689}  & 0.760  & 0.786   \\
        HMC & \textbf{1.472}  & 2.867  & 3.335  & \textbf{1.990}  & 8.211  & 12.777  & \textbf{2.633}  & 9.397  & 12.423  & \textbf{2.226}  & 5.079  & 3.920 \\
        
\toprule
        ~ & \multicolumn{3}{c}{Driver5} & \multicolumn{3}{c}{Driver6} & \multicolumn{3}{c}{Driver7} & \multicolumn{3}{c}{Average}\\
        ~ & HOCD & DCCD & FACD & HOCD & DCCD & FACD & HOCD & DCCD & FACD & HOCD & DCCD & FACD \\
\midrule
        Safety & \textbf{9.838}  & 10.993  & 12.089  & \textbf{9.953} & 11.854  & 11.682  & \textbf{9.512}  & 10.408  & 10.133  & \textbf{9.789}  & 11.081  & 11.394   \\
        Stability & \textbf{17.707}  & 18.206  & 17.858  & \textbf{17.138}  & 18.331  & 18.938  & \textbf{17.325}  & 17.968  & 18.471  & \textbf{17.498}  & 18.200  & 18.237   \\
        Comfort & 12.060  & 14.070  & \textbf{11.116}  & 11.877  & \textbf{11.350}  & 11.371  & 11.291  & \textbf{11.216}  & 14.021  & \textbf{11.454}  & 12.488  & 12.013   \\
        DPW & \textbf{6.843}  & 14.578  & 26.549  & \textbf{2.933}  & 23.105  & 22.371  & \textbf{4.454}  & 11.681  & 7.666  & \textbf{6.135}  & 16.290  & 19.814   \\
        DCW & \textbf{0.609}  & 0.678  & 0.649  & \textbf{0.695}  & 1.070  & 1.055  & \textbf{0.685}  & 0.764  & 0.865  & \textbf{0.662}  & 0.795  & 0.785   \\
        HMC & \textbf{2.085}  & 4.726  & 10.674  & \textbf{3.978}  & 8.593  & 8.198  & \textbf{2.439}  & 4.974  & 3.150  & \textbf{2.403}  & 6.264  & 7.782   \\

\bottomrule
\end{tabular}
\label{table6}
\vskip -0.1in
\end{table*}

\begin{table}[!t]
    \caption{Results of questionnaire.}
    \centering

    \begin{tabular}{cccccc}
    \toprule
        No. & Evaluation & MD & HOCD & DCCD & FACD \\
    \midrule
       1 & Assistance & 0.000 & 3.000 & 2.857 & 2.714 \\
       2 & Trust & 0.714 & 3.429 & 2.571 & 2.571 \\
       3 & Satisfaction & 1.286 & 3.571 & 2.714 & 2.571 \\
    \bottomrule
    \end{tabular}
\label{table7}
\vskip -0.1in
\end{table}

Table~\ref{table5} compares MD and HOCD across multiple drivers and performance metrics. Bolded values indicate the best outcomes. Notably, six out of seven drivers achieved the highest improvement in the safety metric, with values of 0.978, 1.099, 1.688, 0.864, 1.386, and 2.176, respectively. On average, the relative improvement in safety was the highest at 1.303, followed by a significant reduction in the driver physical workload. Overall, driving performance, as measured by safety, stability, and jerk metrics, improves in our proposed HOCD approach compared to MD. Additionally, HOCD reduces both the physical and cognitive workloads of drivers compared to MD.

Table~\ref{table6} presents a comparison of different cooperative driving approaches ( FACD, DCCD, and HOCD) across multiple drivers and metrics. In the DCCD approach, the comfort metric shows the highest improvement for drivers 1, 6, and 7, while the stability metric exhibits the greatest improvement for driver 1. Our proposed HOCD approach demonstrates the greatest improvement for all drivers in the safety metric, along with the largest reductions in driver physical workload, cognitive workload, and human-machine conflict. On average, HOCD achieves superior performance across safety, stability, and comfort metrics, with values of 9.789, 17.498, and 11.454, respectively, outperforming both DCCD and FACD. Besides, HOCD achieves the most significant average reductions in driver physical workload, cognitive workload, and human-machine conflict, with values of 6.135, 0.662, and 2.403, respectively.


Table~\ref{table7} presents the average scores across the four approaches for each evaluation item. The results show that HOCD, DCCD, and FACD were all perceived as providing effective and supportive assistance, with average scores of 3.000, 2.857, and 2.714, respectively. In terms of trust during the NDRT, HOCD received a higher score of 3.429 compared to DCCD and FACD, suggesting that drivers felt more confident with this approach. This result aligns with the lower driver cognitive workload observed in the objective metrics. Furthermore, HOCD achieved the highest overall satisfaction score of 3.571, indicating stronger driver acceptance.

In summary, these results demonstrate that the proposed HOCD approach not only improves objective metrics—such as driving performance, reduction of driver workload, and mitigation of human–machine conflict—but also enhances subjective perceptions of trust and satisfaction. The consistency between subjective evaluations and objective performance further supports the feasibility and effectiveness of the proposed HOCD approach.

\section{Conclusion}
This paper presents a novel human-oriented cooperative driving (HOCD) approach aimed at fostering human-oriented interaction between the driver and automation. By accounting for driver intention and state, the proposed approach effectively reduces human-machine conflict at both the tactical and operational levels. At the tactical level, an intention-aware trajectory planning method was developed, utilizing intention consistency cost as the core metric to align the trajectory with driver intention. At the operational level, a control authority allocation strategy based on reinforcement learning was introduced, optimizing the policy through a designed reward function to maintain consistency between driver state and authority allocation.
The experimental results validate the effectiveness of HOCD from both qualitative and quantitative perspectives, as well as from subjective and objective evaluations. Specifically, HOCD significantly reduces human-machine conflict (2.403), compared to DCCD (6.264) and FACD (7.782). Moreover, HOCD received the highest overall satisfaction score (3.571), outperforming DCCD (2.714) and FACD (2.571) in subjective evaluations.

This study provides critical support for advancing human-vehicle cooperative driving from the driver's perspective, offering a practical and feasible transition path toward fully autonomous driving. Although the proposed approach demonstrates effectiveness across various driving scenarios, its validation is primarily limited to simulation environments, lacking thorough evaluation under real-world conditions.
In the future, the driver model could be further personalized to accommodate individual preferences and behavioral patterns, including potential cultural differences in driving norms and expectations, by using deep learning methods to model more complex and diverse driving behaviors. In addition, integration with advanced driver intention recognition technologies, such as brain-computer interfaces or physiological signal monitoring, could enhance the naturalness and fluidity of human-vehicle interaction. The proposed approach will also be deployed and evaluated in real-world environments to assess its practical effectiveness and adaptability.

\bibliographystyle{IEEEtran}
\bibliography{ref}

@article{sever2024automated,
  title={Automated driving regulations--where are we now?},
  author={Sever, Tina and Contissa, Giuseppe},
  journal={Transportation Research Interdisciplinary Perspectives},
  volume={24},
  pages={101033},
  year={2024},
  publisher={Elsevier}
}

@article{zhang2024human,
  title={Human-Machine Shared Control for Industrial Vehicles: A Personalized Driver Behavior Recognition and Authority Allocation Scheme},
  author={Zhang, Yang and Lu, Jianwei and Xia, Guang and Khajepour, Amir},
  journal={IEEE Transactions on Intelligent Vehicles},
  year={2024},
  volume={},
  number={},
  pages={1-12},
  doi={10.1109/TIV.2024.3389952},
  publisher={IEEE}
}

@article{3muzahid2024survey,
  title={Survey on Human-Vehicle Interactions and AI Collaboration for Optimal Decision-Making in Automated Driving},
  author={Muzahid, Abu Jafar Md and Zhao, Xiaopeng and Wang, Zhenbo},
  journal={arXiv preprint arXiv:2412.08005},
  year={2024}
}

@article{tan2021human,
  title={Human--machine interaction in intelligent and connected vehicles: A review of status quo, issues, and opportunities},
  author={Tan, Zhengyu and Dai, Ningyi and Su, Yating and Zhang, Ruifo and Li, Yijun and Wu, Di and Li, Shutao},
  journal={IEEE Transactions on Intelligent Transportation Systems},
  volume={23},
  number={9},
  pages={13954--13975},
  year={2021},
  publisher={IEEE}
}

@article{peintner2024design,
  title={How to Design Human-Vehicle Cooperation for Automated Driving: A Review of Use Cases, Concepts, and Interfaces},
  author={Peintner, Jakob and Escher, Bengt and Detjen, Henrik and Manger, Carina and Riener, Andreas},
  journal={Multimodal Technologies and Interaction},
  volume={8},
  number={3},
  pages={16},
  year={2024},
  publisher={MDPI}
}

@article{5marcano2020review,
  title={A review of shared control for automated vehicles: Theory and applications},
  author={Marcano, Mauricio and D{\'\i}az, Sergio and P{\'e}rez, Joshu{\'e} and Irigoyen, Eloy},
  journal={IEEE Transactions on Human-Machine Systems},
  volume={50},
  number={6},
  pages={475--491},
  year={2020},
  publisher={IEEE}
}

@inproceedings{6sheridan1978human,
  title={Human/computer control of undersea teleoperators},
  author={Sheridan, Thomas B and Verplank, William L and Brooks, TL},
  booktitle={NASA. Ames Res. Center The 14th Ann. Conf. on Manual Control},
  year={1978}
}

@article{7yang2021review,
  title={A review of human--machine cooperation in the robotics domain},
  author={Yang, Canjun and Zhu, Yuanchao and Chen, Yanhu},
  journal={IEEE Transactions on Human-Machine Systems},
  volume={52},
  number={1},
  pages={12--25},
  year={2021},
  publisher={IEEE}
}

@article{8wang2020decision,
  title={Decision-making in driver-automation shared control: A review and perspectives},
  author={Wang, Wenshuo and Na, Xiaoxiang and Cao, Dongpu and Gong, Jianwei and Xi, Junqiang and Xing, Yang and Wang, Fei-Yue},
  journal={IEEE/CAA Journal of Automatica Sinica},
  volume={7},
  number={5},
  pages={1289--1307},
  year={2020},
  publisher={IEEE}
}

@article{9fang2023human,
  title={A human-machine shared control framework considering time-varying driver characteristics},
  author={Fang, Zhenwu and Wang, Jinxiang and Wang, Zejiang and Liang, Jinhao and Liu, Yahui and Yin, Guodong},
  journal={IEEE Transactions on Intelligent Vehicles},
  volume={8},
  number={7},
  pages={3826--3838},
  year={2023},
  publisher={IEEE}
}

@article{10lu2022shared,
  title={A shared control design for steering assistance system considering driver behaviors},
  author={Lu, Yanbo and Liang, Jinhao and Yin, Guodong and Xu, Liwei and Wu, Jian and Feng, Jiwei and Wang, Faan},
  journal={IEEE Transactions on Intelligent Vehicles},
  volume={8},
  number={1},
  pages={900--911},
  year={2022},
  publisher={IEEE}
}

@article{11fang2023servo,
  title={Servo-Level Human--Machine Shared Control Flexible Strategy Based on Driving Ability, Status and Regionalized Environmental Risk},
  author={Fang, Liu and Tianhe, Zhu and Weixing, Su},
  journal={IEEE Transactions on Transportation Electrification},
  volume={9},
  number={3},
  pages={4418--4436},
  year={2023},
  publisher={IEEE}
}

@article{12fang2024human,
  title={Human--machine shared control for path following considering driver fatigue characteristics},
  author={Fang, Zhenwu and Wang, Jinxiang and Wang, Zejiang and Chen, Jinxin and Yin, Guodong and Zhang, Hui},
  journal={IEEE Transactions on Intelligent Transportation Systems},
  year={2024},
  publisher={IEEE}
}

@book{14abbink2010neuromuscular,
  title={Neuromuscular analysis as a guideline in designing shared control},
  author={Abbink, David A and Mulder, Mark},
  year={2010},
  publisher={INTECH Open Access Publisher}
}

@article{15mars2014analysis,
  title={Analysis of human-machine cooperation when driving with different degrees of haptic shared control},
  author={Mars, Franck and Deroo, Mathieu and Hoc, Jean-Michel},
  journal={IEEE Transactions on Haptics},
  volume={7},
  number={3},
  pages={324--333},
  year={2014},
  publisher={IEEE}
}

@article{16abbink2012haptic,
  title={Haptic shared control: smoothly shifting control authority?},
  author={Abbink, David A and Mulder, Mark and Boer, Erwin R},
  journal={Cognition, Technology \& Work},
  volume={14},
  pages={19--28},
  year={2012},
  publisher={Springer}
}

@article{17li2018shared,
  title={Shared control driver assistance system based on driving intention and situation assessment},
  author={Li, Mingjun and Cao, Haotian and Song, Xiaolin and Huang, Yanjun and Wang, Jianqiang and Huang, Zhi},
  journal={IEEE Transactions on Industrial Informatics},
  volume={14},
  number={11},
  pages={4982--4994},
  year={2018},
  publisher={IEEE}
}

@article{18jiang2017shared,
  title={Shared-control for a rear-wheel drive car: Dynamic environments and disturbance rejection},
  author={Jiang, Jingjing and Astolfi, Alessandro},
  journal={IEEE Transactions on Human-Machine Systems},
  volume={47},
  number={5},
  pages={723--734},
  year={2017},
  publisher={IEEE}
}

@article{19li2021shared,
  title={Shared steering control for human--machine co-driving system with multiple factors},
  author={Li, Xueyun and Wang, Yiping},
  journal={Applied Mathematical Modelling},
  volume={100},
  pages={471--490},
  year={2021},
  publisher={Elsevier}
}

@article{20nguyen2017sensor,
  title={Sensor reduction for driver-automation shared steering control via an adaptive authority allocation strategy},
  author={Nguyen, Anh-Tu and Sentouh, Chouki and Popieul, Jean-Christophe},
  journal={IEEE/ASME Transactions on Mechatronics},
  volume={23},
  number={1},
  pages={5--16},
  year={2017},
  publisher={IEEE}
}

@article{23_2022adaptive,
  title={Adaptive authority allocation approach for shared steering control system},
  author={Li, Xueyun and Wang, Yiping and Su, Chuqi and Gong, Xinle and Huang, Jin and Yang, Dengke},
  journal={IEEE Transactions on Intelligent Transportation Systems},
  volume={23},
  number={10},
  pages={19428--19439},
  year={2022},
  publisher={IEEE}
}

@article{24liu2022human,
  title={Human-oriented online driving authority optimization for driver-automation shared steering control},
  author={Liu, Jun and Dai, Qikun and Guo, Hongyan and Guo, Jingzheng and Chen, Hong},
  journal={IEEE Transactions on Intelligent Vehicles},
  volume={7},
  number={4},
  pages={863--872},
  year={2022},
  publisher={IEEE}
}

@article{25dai2023bargaining,
  title={A bargaining game-based human--machine shared driving control authority allocation strategy},
  author={Dai, Changhua and Zong, Changfu and Zhang, Dong and Hua, Min and Zheng, Hongyu and Chuyo, Kaku},
  journal={IEEE Transactions on Intelligent Transportation Systems},
  volume={24},
  number={10},
  pages={10572--10586},
  year={2023},
  publisher={IEEE}
}

@article{26guo2023game,
  title={Game-theoretic Human-Machine Shared Steering Control Strategy Under Extreme Conditions},
  author={Guo, Hongyan and Shi, Wanqing and Liu, Jun and Guo, Jingzheng and Meng, Qingyu and Cao, Dongpu and Chen, Hong},
  journal={IEEE Transactions on Intelligent Vehicles},
  year={2024},
  volume={9},
  number={1},
  pages={2766-2779},
  publisher={IEEE}
}

@article{27elbanhawi2014sampling,
  title={Sampling-based robot motion planning: A review},
  author={Elbanhawi, Mohamed and Simic, Milan},
  journal={IEEE Access},
  volume={2},
  pages={56--77},
  year={2014},
  publisher={IEEE}
}

@article{28kavraki1996probabilistic,
  title={Probabilistic roadmaps for path planning in high-dimensional configuration spaces},
  author={Kavraki, Lydia E and Svestka, Petr and Latombe, J-C and Overmars, Mark H},
  journal={IEEE Transactions on Robotics and Automation},
  volume={12},
  number={4},
  pages={566--580},
  year={1996},
  publisher={IEEE}
}

@article{29lavalle2001randomized,
  title={Randomized kinodynamic planning},
  author={LaValle, Steven M and Kuffner Jr, James J},
  journal={The International Journal of Robotics Research},
  volume={20},
  number={5},
  pages={378--400},
  year={2001},
  publisher={SAGE Publications}
}

@article{30lim2018hierarchical,
  title={Hierarchical trajectory planning of an autonomous car based on the integration of a sampling and an optimization method},
  author={Lim, Wonteak and Lee, Seongjin and Sunwoo, Myoungho and Jo, Kichun},
  journal={IEEE Transactions on Intelligent Transportation Systems},
  volume={19},
  number={2},
  pages={613--626},
  year={2018},
  publisher={IEEE}
}

@article{31li2022autonomous,
  title={Autonomous driving on curvy roads without reliance on frenet frame: A cartesian-based trajectory planning method},
  author={Li, Bai and Ouyang, Yakun and Li, Li and Zhang, Youmin},
  journal={IEEE Transactions on Intelligent Transportation Systems},
  volume={23},
  number={9},
  pages={15729--15741},
  year={2022},
  publisher={IEEE}
}

@article{32guo2022down,
  title={Down-sized initialization for optimization-based unstructured trajectory planning by only optimizing critical variables},
  author={Guo, Yuqing and Yao, Danya and Li, Bai and Gao, Haichuan and Li, Li},
  journal={IEEE Transactions on Intelligent Vehicles},
  volume={8},
  number={1},
  pages={709--720},
  year={2022},
  publisher={IEEE}
}

@article{33song2023review,
  title={A Review of the Motion Planning and Control Methods for Automated Vehicles},
  author={Song, Xiaohua and Gao, Huihui and Ding, Tian and Gu, Yunfeng and Liu, Jing and Tian, Kun},
  journal={Sensors},
  volume={23},
  number={13},
  pages={6140},
  year={2023},
  publisher={MDPI}
}

@inproceedings{34lee2012unified,
  title={A unified framework of the automated lane centering/changing control for motion smoothness adaptation},
  author={Lee, Jin-Woo and Litkouhi, Bakhtiar},
  booktitle={15th International Conference on Intelligent Transportation Systems},
  pages={282--287},
  year={2012},
}

@article{35yang20132d,
  title={2D Dubins path in environments with obstacle},
  author={Yang, Dongxiao and Li, Didong and Sun, Huafei},
  journal={Mathematical Problems in Engineering},
  volume={2013},
  number={1},
  pages={291372},
  year={2013},
  publisher={Wiley Online Library}
}

@inproceedings{36qian2016motion,
  title={Motion planning for urban autonomous driving using B{\'e}zier curves and MPC},
  author={Qian, Xiangjun and Navarro, I{\~n}aki and de La Fortelle, Arnaud and Moutarde, Fabien},
  booktitle={19th International Conference on Intelligent Transportation Systems (ITSC)},
  pages={826--833},
  year={2016},
}

@article{37maekawa2010curvature,
  title={Curvature continuous path generation for autonomous vehicle using B-spline curves},
  author={Maekawa, Takashi and Noda, Tetsuya and Tamura, Shigefumi and Ozaki, Tomonori and Machida, Ken-ichiro},
  journal={Computer-Aided Design},
  volume={42},
  number={4},
  pages={350--359},
  year={2010},
  publisher={Elsevier}
}

@article{38hwang2003fast,
  title={A fast path planning by path graph optimization},
  author={Hwang, Joo Young and Kim, Jun Song and Lim, Sang Seok and Park, Kyu Ho},
  journal={IEEE Transactions on Systems, Man, and Cybernetics-Part a: Systems and Humans},
  volume={33},
  number={1},
  pages={121--129},
  year={2003},
  publisher={IEEE}
}

@article{39hart1968formal,
  title={A formal basis for the heuristic determination of minimum cost paths},
  author={Hart, Peter E and Nilsson, Nils J and Raphael, Bertram},
  journal={IEEE Transactions on Systems Science and Cybernetics},
  volume={4},
  number={2},
  pages={100--107},
  year={1968},
  publisher={IEEE}
}

@misc{40kurzer2016path,
  title={Path planning in unstructured environments: A real-time hybrid a* implementation for fast and deterministic path generation for the kth research concept vehicle},
  author={Kurzer, Karl},
  year={2016}
}

@inproceedings{41stentz1994optimal,
  title={Optimal and efficient path planning for partially-known environments},
  author={Stentz, Anthony},
  booktitle={Proceedings of the 1994 IEEE international conference on robotics and automation},
  pages={3310--3317},
  year={1994},
}

@article{42wang2016human,
  title={Human-centered feed-forward control of a vehicle steering system based on a driver's path-following characteristics},
  author={Wang, Wenshuo and Xi, Junqiang and Liu, Chang and Li, Xiaohan},
  journal={IEEE Transactions on Intelligent Transportation Systems},
  volume={18},
  number={6},
  pages={1440--1453},
  year={2016},
  publisher={IEEE}
}

@article{43bolia2014driver,
  title={Driver steering model for closed-loop steering function analysis},
  author={Bolia, Pratiksh and Weiskircher, Thomas and M{\"u}ller, Steffen},
  journal={Vehicle System Dynamics},
  volume={52},
  number={sup1},
  pages={16--30},
  year={2014},
  publisher={Taylor \& Francis}
}

@article{44powell2011relationships,
  title={Relationships between lane change performance and open-loop vehicle handling metrics},
  author={Powell, Robert and Ayalew, Beshah and Law, E Harry},
  journal={International Journal of Vehicle Design},
  volume={56},
  number={1-4},
  pages={19--33},
  year={2011},
  publisher={Inderscience Publishers}
}

@article{45deng2022shared,
  title={Shared control for intelligent vehicle based on handling inverse dynamics and driving intention},
  author={Deng, Huifan and Zhao, Youqun and Feng, Shilin and Wang, Qiuwei and Lin, Fen},
  journal={IEEE Transactions on Vehicular Technology},
  volume={71},
  number={3},
  pages={2706--2720},
  year={2022},
  publisher={IEEE}
}

@article{46jin2025impact,
  title={Impact of non-driving related task types, request modalities, and automation on driver takeover: A meta-analysis},
  author={Jin, Lisheng and Liu, Xingchen and Guo, Baicang and Han, Zhuotong and Wang, Yinlin and Cao, Yuan and Yang, Xiao and Shi, Jian},
  journal={Safety Science},
  volume={181},
  pages={106704},
  year={2025},
  publisher={Elsevier}
}

@article{47li2009synthetic,
  title={Synthetic assessment of cognitive load in human-machine interaction process},
  author={Li, Jin-Bo and Xu, Bai-Hua},
  journal={Acta Psychologica Sinica},
  volume={41},
  number={01},
  pages={35},
  year={2009}
}

@article{48yan2024human,
  title={Human-Vehicle Shared Steering Control for Obstacle Avoidance: A Reference-Free Approach With Reinforcement Learning},
  author={Yan, Liang and Wu, Xiaodong and Wei, Chongfeng and Zhao, Sheng},
  journal={IEEE Transactions on Intelligent Transportation Systems},
  year={2024},
  pages={17888--17901},
  volume={25},
  number={11},  
  publisher={IEEE}
}

@article{49zhang2021driving,
  title={Driving authority allocation strategy based on driving authority real-time allocation domain},
  author={Zhang, Ziyu and Wang, Chunyan and Zhao, Wanzhong and Xu, Can and Chen, Guoping},
  journal={IEEE Transactions on Intelligent Transportation Systems},
  volume={23},
  number={7},
  pages={8528--8543},
  year={2021},
  publisher={IEEE}
}

@article{50sentouh2018driver,
  title={Driver-automation cooperation oriented approach for shared control of lane keeping assist systems},
  author={Sentouh, Chouki and Nguyen, Anh-Tu and Benloucif, Mohamed Amir and Popieul, Jean-Christophe},
  journal={IEEE Transactions on Control Systems Technology},
  volume={27},
  number={5},
  pages={1962--1978},
  year={2018},
  publisher={IEEE}
}

@book{52guo2008,
  author    = {Konghui Guo},
  title     = {Vehicle Handling Dynamics},
  publisher = {Phoenix Science Press},
  year      = {2011}
}

@article{53falcone2007predictive,
  title={Predictive active steering control for autonomous vehicle systems},
  author={Falcone, Paolo and Borrelli, Francesco and Asgari, Jahan and Tseng, Hongtei Eric and Hrovat, Davor},
  journal={IEEE Transactions on Control Systems Technology},
  volume={15},
  number={3},
  pages={566--580},
  year={2007},
  publisher={IEEE}
}

@inproceedings{55werling2010optimal,
  title={Optimal trajectory generation for dynamic street scenarios in a frenet frame},
  author={Werling, Moritz and Ziegler, Julius and Kammel, S{\"o}ren and Thrun, Sebastian},
  booktitle={IEEE International Conference on Robotics and Automation},
  pages={987--993},
  year={2010},
}

@inproceedings{56dosovitskiy2017carla,
  title={CARLA: An open urban driving simulator},
  author={Dosovitskiy, Alexey and Ros, German and Codevilla, Felipe and Lopez, Antonio and Koltun, Vladlen},
  booktitle={Conference on Robot Learning},
  pages={1--16},
  year={2017},
  organization={PMLR}
}

@inproceedings{sac_haarnoja2018soft,
  title={Soft actor-critic: Off-policy maximum entropy deep reinforcement learning with a stochastic actor},
  author={Haarnoja, Tuomas and Zhou, Aurick and Abbeel, Pieter and Levine, Sergey},
  booktitle={International Conference on Machine Learning},
  pages={1861--1870},
  year={2018},
  organization={PMLR}
}

@article{ppo_schulman2017proximal,
  title={Proximal policy optimization algorithms},
  author={Schulman, John and Wolski, Filip and Dhariwal, Prafulla and Radford, Alec and Klimov, Oleg},
  journal={arXiv preprint arXiv:1707.06347},
  year={2017}
}

@article{ddpg_lillicrap2015continuous,
  title={Continuous control with deep reinforcement learning},
  author={Lillicrap, TP},
  journal={arXiv preprint arXiv:1509.02971},
  year={2015}
}

@article{rl-xie2022coordination,
  title={Coordination control strategy for human-machine cooperative steering of intelligent vehicles: A reinforcement learning approach},
  author={Xie, Ju and Xu, Xing and Wang, Feng and Liu, Zhenyu and Chen, Long},
  journal={IEEE Transactions on Intelligent Transportation Systems},
  volume={23},
  number={11},
  pages={21163--21177},
  year={2022},
  publisher={IEEE}
}

@article{rl-wang2024human,
  title={Human-machine Authority Allocation in Indirect Cooperative Shared Steering Control with TD3 Reinforcement Learning},
  author={Wang, Hongbo and Feng, Lizhao and Zhang, Yuhong and Zhou, Juntao and Du, Haiping},
  journal={IEEE Transactions on Vehicular Technology},
  year={2024},
  volume={73},
  number={6},
  pages={7576-7588},
  publisher={IEEE}
}

@inproceedings{shen2021distributed,
  title={Distributed nonlinear model predictive control for heterogeneous vehicle platoons under uncertainty},
  author={Shen, Dan and Yin, Jianhua and Du, Xiaoping and Li, Lingxi},
  booktitle={2021 IEEE International Intelligent Transportation Systems Conference (ITSC)},
  pages={3596--3603},
  year={2021},
  organization={IEEE}
}

@article{yin2022distributed,
  title={Distributed stochastic model predictive control with Taguchi’s robustness for vehicle platooning},
  author={Yin, Jianhua and Shen, Dan and Du, Xiaoping and Li, Lingxi},
  journal={IEEE Transactions on Intelligent Transportation Systems},
  volume={23},
  number={9},
  pages={15967--15979},
  year={2022},
  publisher={IEEE}
}

\vfill

\end{document}